\documentclass[10pt]{article} 
\usepackage[preprint]{tmlr}

\usepackage{amsmath,amsfonts,bm}

\def\eqref#1{equation~\ref{#1}}

\def\1{\bm{1}}

\DeclareMathAlphabet{\mathsfit}{\encodingdefault}{\sfdefault}{m}{sl}
\SetMathAlphabet{\mathsfit}{bold}{\encodingdefault}{\sfdefault}{bx}{n}

\usepackage{hyperref}
\usepackage{url}
\usepackage{graphicx}
\usepackage{subcaption}
\usepackage[edges]{forest}
\usepackage{adjustbox}
\usepackage{booktabs}    
\usepackage{multirow}    
\usepackage{rotating}    
\usepackage{makecell}    
\usepackage{tabularx}
\usepackage{amsmath}
\usepackage{tikz}
\tikzstyle{leaf}=[ 
    minimum height=1.5em,
    text=black,
    align=left,
    font=\normalsize,
    inner xsep=2pt,
    inner ysep=4pt,
]

\usepackage{pifont}      

\title{\Large Spatial Reasoning in Multimodal Large Language Models: A Survey of Tasks, Benchmarks and Methods}
\author{\name Weichen Liu \email weichenliu@pitt.edu \\
      \addr University of Pittsburgh
      \AND
      \name Qiyao Xue \email qix63@pitt.edu\\
      \addr University of Pittsburgh
      \AND
      \name Haoming Wang \email hw.wang@pitt.edu\\
      \addr University of Pittsburgh
      \AND
      \name Xiangyu Yin \email eric.yin@pitt.edu\\
      \addr University of Pittsburgh
      \AND
      \name Boyuan Yang \email by.yang@pitt.edu\\
      \addr University of Pittsburgh
      \AND
      \name Wei Gao \email weigao@pitt.edu\\
      \addr University of Pittsburgh}

\def\month{MM}  
\def\year{YYYY} 
\def\openreview{\url{https://openreview.net/forum?id=XXXX}} 
\begin{document}
\maketitle

\begin{abstract}
    Spatial reasoning, which requires ability to perceive and manipulate spatial relationships in the 3D world, is a fundamental aspect of human intelligence, yet remains a persistent challenge for Multimodal large language models (MLLMs). While existing surveys often categorize recent progress based on input modality (e.g., text, image, video, or 3D), we argue that spatial ability is not solely determined by the input format. Instead, our survey introduces a taxonomy that organizes spatial intelligence from cognitive aspect and divides tasks in terms of reasoning complexity, linking them to several cognitive functions. We map existing benchmarks across text-only, vision–language, and embodied settings onto this taxonomy, and review evaluation metrics and methodologies for assessing spatial reasoning ability. This cognitive perspective enables more principled cross-task comparisons and reveals critical gaps between current model capabilities and human-like reasoning. In addition, we analyze methods for improving spatial ability, spanning both training-based and reasoning-based approaches. This dual-perspective analysis clarifies their respective strengths, uncovers complementary mechanisms. By surveying tasks, benchmarks, and recent advances, we aim to provide new researchers with a comprehensive understanding of the field and actionable directions for future research.
\end{abstract}

\section{Introduction}
The development of Large language model (LLM) represents a significant milestone in artificial intelligence, showcasing unprecedented capabilities in comprehending, reasoning over, and generating human-like natural language. These models, built upon deep learning architectures like the Transformer\citep{2402.06196, 10.5555/3295222.3295349}, are pre-trained on vast amounts of linguistic corpora, enabling them to perform a wide array of language-centric tasks, from translation and summarization to complex reasoning\citep{electronics13245040, 2310.12321}. The introduction of models such as GPT-3 has showcased their potential in few-shot learning\citep{2005.14165}, where they can adapt to new tasks with minimal examples. These advances have established \textbf{language intelligence} as a cornerstone of contemporary artificial intelligence research, demonstrating scalable generalization across a wide range of linguistic tasks.

Building on LLM's linguistic foundation, researchers seek to move towards visual perception and understanding. Vision-Language Models (VLMs) represent an ongoing attempt to bridge perception and language by coupling visual encoders\citep{dosovitskiy2021an, liu2021swintransformerhierarchicalvision, radford2021learningtransferablevisualmodels, liu2022convnet2020s} with pretrained LLM backbones. However, a critical aspect of human-like intelligence that remains a significant challenge for these models is \textbf{spatial intelligence}, which is the ability to perceive, understand, and reason about the spatial relationships between objects, their orientation, and their movement in both real and imaginary spaces. While humans navigate and interact with the three-dimensional world easily, equipping LLMs and VLMs with a comparable level of spatial understanding is an ongoing frontier in AI research.

\begin{figure}[t]
    \centering
    \begin{subfigure}{1\linewidth}
        \centering
        \includegraphics[width=\linewidth]{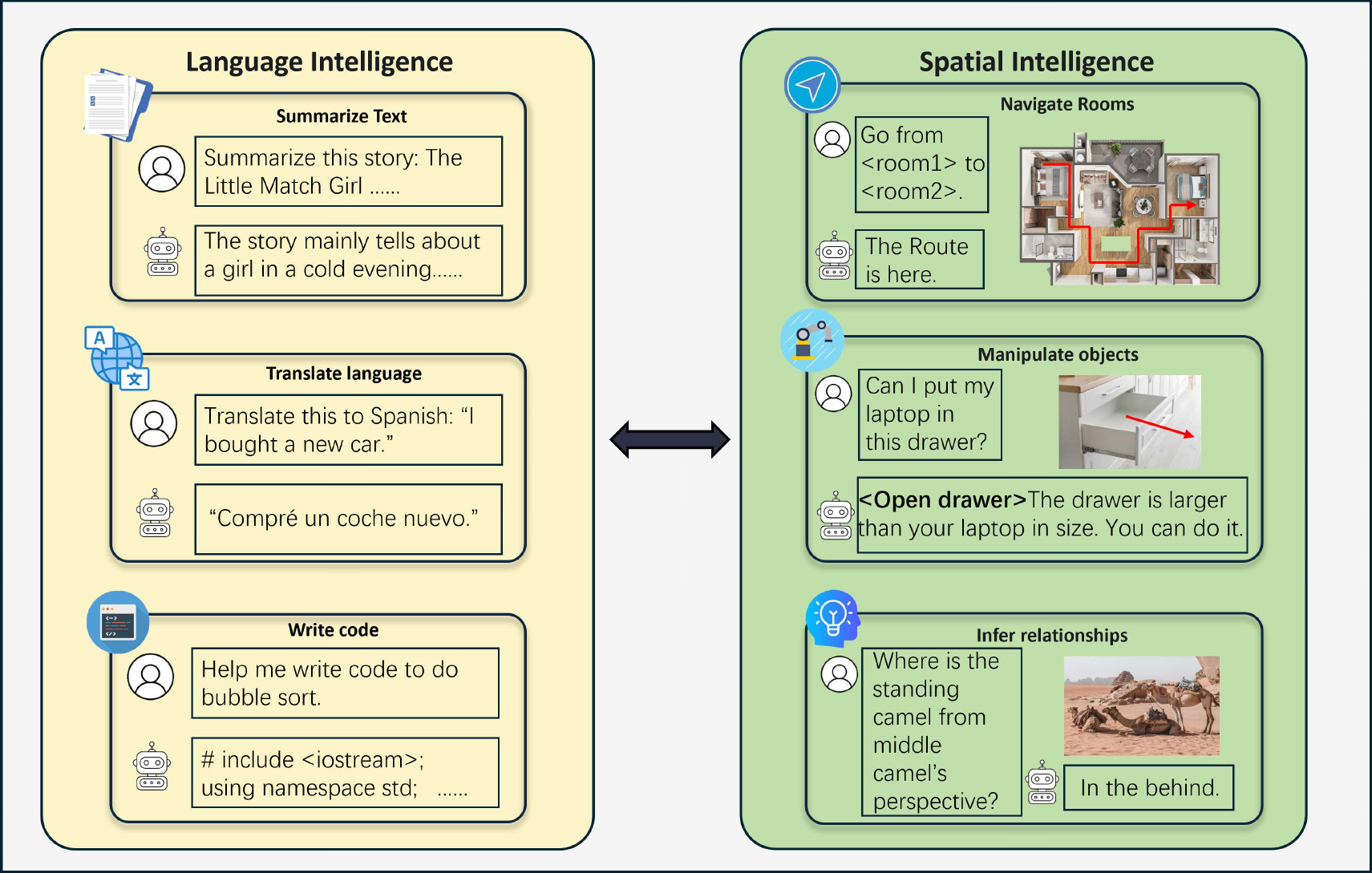}
        \caption{Different tasks in language and spatial intelligence}
        \label{fig:lanvsspatial}
    \end{subfigure}

    \vspace{2em} 

    \begin{subfigure}{1\linewidth}
        \centering
        \includegraphics[width=\linewidth]{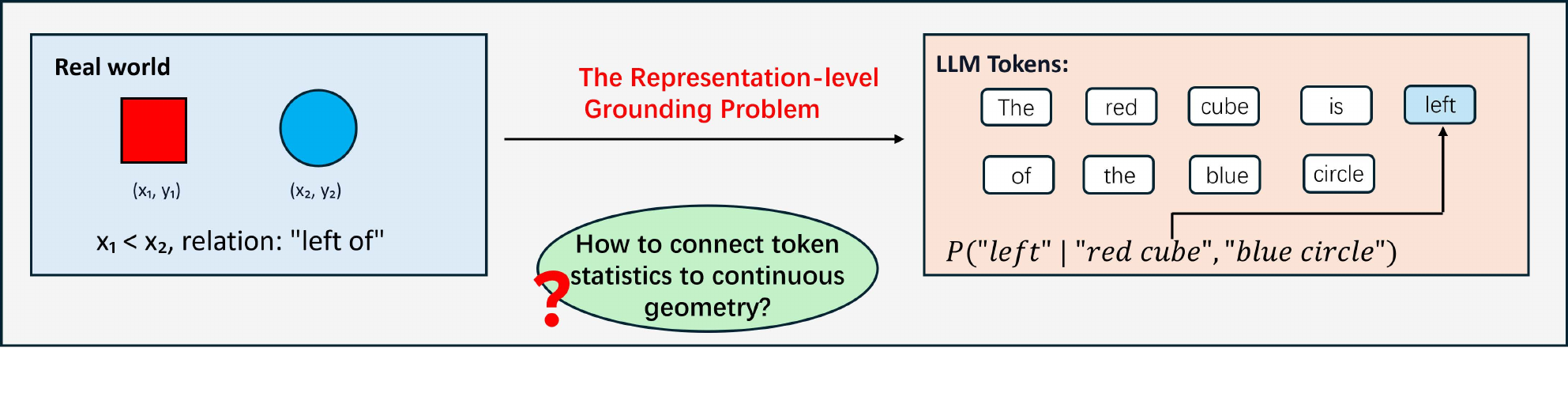}
        \caption{The Representational Mismatch}
        \label{fig:difficulty}
    \end{subfigure}

    \caption{The gap between language intelligence and spatial intelligence for MLLMs: 
    (a) Language tasks rely on discrete and sequential token processing, while spatial tasks require grounded reasoning in continuous 3D space.
    (b) This mismatch reflects the representation-level grounding problem—MLLMs model statistical co-occurrence rather than true geometric relations.}
    \label{fig:lanvsspatial-difficulty}
\end{figure}

The gap between language intelligence and spatial intelligence manifested explicitly across distinct task categories and implicitly through different human mental representations as shown in Figure~\ref{fig:lanvsspatial-difficulty}. LLMs demonstrate strong proficiency in language intelligence, operating over linguistic data such as text and code. In contrast, spatial intelligence involves reasoning within a grounded, 3D environment, enabling capabilities such as robot navigation and object manipulation. The underlying cause of this disparity lies in a fundamental representational mismatch: whereas the physical world is characterized by continuous geometric structures, LLMs encode and interpret information as discrete, sequential tokens. They learn spatial concepts not as geometric principles, but as statistical co-occurrences bias in vast datasets of text and images. For example, they learn that the words ``left of'' statistically appear between ``cube'' and ``circle'' without a true geometric understanding of the relationship. 

This fundamental distinction between linguistic and spatial intelligence is deeply rooted in both cognitive science and neural mechanisms. From a cognitive science perspective, humans reason about space not through linguistic expression alone, but through mental models, which is an internal analog representations preserving geometric and topological relations among objects\citep{10.5555/7909, tversky1993cognitive}. These mental models allow people to mentally simulate transformations such as rotation, translation, and perspective change\citep{BYRNE1989564}. In contrast, language compresses these continuous relationships into discrete categorical tokens (e.g., ``left'', ``behind'', ``on top of''), which describe qualitative spatial relations rather than quantitatively encoding geometry\citep{DBLP:reference/fai/CohnR08}. The evidence of neural mechanisms further underscores this difference. The hippocampal–entorhinal circuit is known to support spatial cognition through ``place cells'' and ``grid cells'', which encode allocentric maps and metric structure of the environment \citep{OKEEFE1971171, hafting2005microstructure, stensola2016grid}. These continuous neural codes form an internal coordinate system enabling path integration, location tracking, and mental navigation through space \citep{basu2024neural}. In contrast, the representational substrate of language in the cortex is largely sequential and discrete, optimized for linguistic composition and token prediction, not for metric spatial computation.

This mismatch between analog spatial coding and discrete linguistic encoding represents the classic representation-level grounding problem \citep{harnad1990symbol}. LLMs are lack of an internal map of space that would allow them to mentally model or adjust these spatial relationships as humans do. Even for VLMs, though grounded in visual perception, typically remain constrained to 2D or projective representations and lack deeper 3D spatial mental modeling. Bridging this gap remains a formidable challenge, our proposed cognitive function taxonomy is specifically designed to reveal and systematically organize these deficiencies within current model capabilities.

\vspace{1em}
\begin{figure}[h]   
  \centering
  \includegraphics[width=1\linewidth]{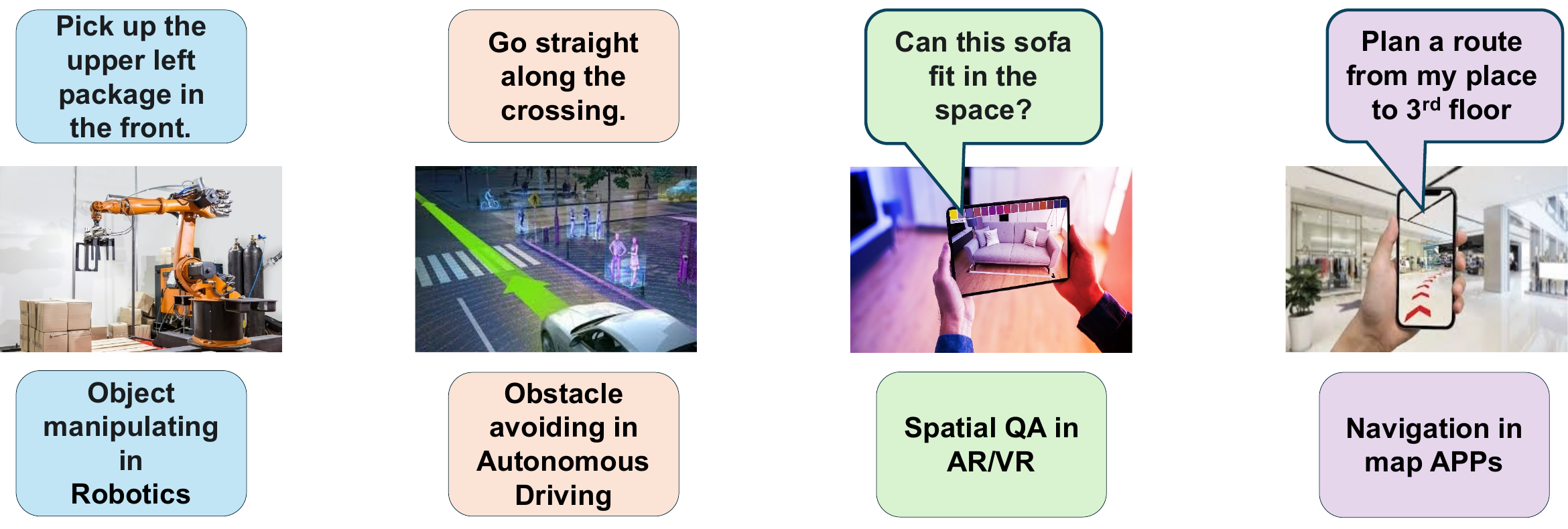}
  \caption{Spatial tasks for different application domains}
  \label{fig:applications}
\end{figure}

\noindent\textbf{Motivation and application:} The advancement of spatial intelligence in LLMs and VLMs is crucial for enabling modern MLLMs to evolve towards comprehensive world models capable of understanding and interacting with the real world. While models have shown remarkable progress in language-centric tasks\citep{2005.14165, wei2023chainofthoughtpromptingelicitsreasoning}, their understanding of spatial relationships remains a significant hurdle\citep{zhang2025mllmsstrugglespatialunderstanding, han2025largelanguagemodelsintegrate}. This limitation becomes critical when considering the vast range of embodied and spatial perception-grounded applications, driven tasks that inherently require coherent spatial reasoning and continuous interaction with the physical world. As illustrated in \ref{fig:applications}, robust spatial intelligence is significant for the successful deployment of MLLMs in a multitude of high-impact domains:

\begin{itemize}
    \item Robotics: For a robot to be truly helpful in unstructured environments, it must be able to reason about the geometric relations between objects and locations\citep{robo}. This includes tasks like grasping objects, avoiding obstacles, and navigating through cluttered spaces, all of which demand a sophisticated understanding of spatial dynamics.
    \item Autonomous Driving: The safety and reliability of self-driving cars are fundamentally dependent on their ability to perceive and reason about their spatial surroundings. This involves interpreting sensor data to understand the positions and trajectories of other vehicles, pedestrians, and road signs to make critical, real-time decisions\citep{autom}.
    \item Augmented and Virtual Reality (AR/VR): In AR and VR, the user's sense of immersion and the utility of the application are directly tied to the system's ability to comprehend and manipulate spatial information\citep{Bozkir_2024, park2024integratinglargelanguagemodels}. For AR, this means accurately overlaying digital information onto the real world, such as navigation instructions in a complex building. In VR, it enables the creation of realistic, interactive virtual environments for training, design, and entertainment.
    \item Navigation: Whether for a person using an AR--enhanced mapping application or for an autonomous robot delivering supplies in a warehouse, the ability to understand and follow spatial instructions is paramount\citep{lin2025advancesembodiednavigationusing}. This requires the model to interpret and act upon language that describes routes, locations, and the relative positions of objects.
\end{itemize}

By endowing MLLMs with strong spatial abilities, we can unlock their potential to move beyond the digital realm and into dynamic, physical environments, paving the way for more capable and safer AI systems.

\begin{figure*}[th!]
    \centering
    \resizebox{\textwidth}{!}{
        \begin{forest}
            forked edges,
            for tree={
                grow=east,
                reversed=true,
                anchor=base west,
                parent anchor=east,
                child anchor=west,
                base=left,
                font=\large,
                rectangle,
                draw=black,
                rounded corners,
                align=left,
                minimum width=4em,
                edge+={gray, line width=1pt},
                s sep=3pt,
                inner xsep=2pt,
                inner ysep=3pt,
                line width=0.8pt,
                ver/.style={rotate=90, child anchor=north, parent anchor=south, anchor=center},
            },
            where level=1{text width=8em,font=\normalsize,}{},
            where level=2{text width=14em,font=\normalsize,}{},
            where level=3{text width=17em,font=\normalsize,}{},
            where level=4{text width=11em,font=\normalsize,}{},
            where level=5{text width=6.5em,font=\normalsize,}{},
            [
                Spatial Reasoning in Multi-modal Large Language Models, ver, fill=gray!20
                [
                    Taxonomy of \\ spatial reasoning \\ tasks \S ~\ref{taxonomy label}, fill=blue!20
                    [
                        Categorize by \\cognitive functions \S ~\ref{cognitive function categorization}, fill=blue!20
                            [Intrinsic – Qualitative – Static, fill=blue!20 [MindCube\citep{yin2025spatialmentalmodelinglimited}{\,,}
                        Super-CLEVR-3D \citep{wang20233d}{\,,}
                        \\
                        Text2Shape Dataset\citep{chen2018text2shape}, leaf, text width=33em, edge=black, fill=blue!20]]
                            [Extrinsic – Qualitative – Static, fill=blue!20 [SPARTQA \citep{mirzaee-etal-2021-spartqa}{\,,} 
                        SpatialEval(VQA) \citep{wang2024spatial}, leaf, text width=33em, edge=black, fill=blue!20]]
                            [Quantitative – Static, fill=blue!20 [Q-Spatial Bench\citep{liao2024reasoningpathsreferenceobjects}{\,,} 
                        ScanRefer\citep{chen2020scanrefer}, leaf, text width=33em, edge=black, fill=blue!20]]
                            [Extrinsic – Qualitative – Dynamic , fill=blue!20 [MindCube\citep{yin2025spatialmentalmodelinglimited}{\,,} 
                        STARE \citep{li2025unfoldingspatialcognitionevaluating}, leaf, text width=33em, edge=black, fill=blue!20]]
                            [Intrinsic – Qualitative – Dynamic, fill=blue!20 [VSI-bench \citep{yang2024think}{\,,} 
                        SQA3D\citep{ma2022sqa3d}{\,,}
                        \\
                        M3DBench\citep{li2023m3dbench}{\,,}
                        3DSRBench \citep{ma20243dsrbench}, leaf, text width=33em, edge=black, fill=blue!20]]
                    ]
                    [
                        Levels of  Reasoning \\Complexity \S ~\ref{reasoning complexity level}, fill=blue!20
                        [Level 1: Direct perception, fill=blue!20[COCO\citep{lin2015microsoftcococommonobjects}{\,,} 
                        QVA dataset \citep{agrawal2016vqavisualquestionanswering}, leaf, text width=33em, edge=black, fill=blue!20]]
                        [Level 2: Single-step inference, fill=blue!20[Super-CLEVR-3D\citep{wang20233d}{\,,} 
                        6dof\_spatialbench \citep{sofar25}, leaf, text width=33em, edge=black, fill=blue!20]]
                        [Level 3: Multi-step chaining, fill=blue!20[3DMV-VQA\citep{hong20233d}{\,,} 
                        VSI-bench \citep{yang2024think}, leaf, text width=33em, edge=black, fill=blue!20]]
                        [Level 4: Advanced synthetic problems, fill=blue!20[MindCube\citep{yin2025spatialmentalmodelinglimited}{\,,} 
                        OmniSpatial \citep{omnispatial25}, leaf, text width=33em, edge=black, fill=blue!20]]
                    ]
                ]
                [
                    Benchmarks \\and evaluation \\metrics \S ~\ref{b and em}, fill=red!20
                    [
                        Existing benchmarks \S ~\ref{eb}, fill=red!20
                        [Text-only Benchmarks , fill=red!20[SpartQA\citep{mirzaee-etal-2021-spartqa}{\,,} 
                        SpatialEval(TQA) \citep{wang2024spatial}{\,,}
                        \\
                        BaBi(task 17-19) \citep{DBLP:journals/corr/WestonBCM15}
                        StepGame \citep{stepGame2022shi}, leaf, text width=33em, edge=black, fill=red!20]]
                        [Image/video Benchmarks , fill=red!20[SpatialEval(VQA) \citep{wang2024spatial}{\,,} 
                        Q-Spatial Bench\citep{liao2024reasoningpathsreferenceobjects}
                        \\
                        VSI-bench \citep{yang2024think}{\,,}
                        MindCube\citep{yin2025spatialmentalmodelinglimited}{\,,}
                        \\
                        EmbSpatial-Bench \citep{du-etal-2024-embspatial}{\,,}
                        ViewSpatial-Bench \citep{li2025viewspatialbenchevaluatingmultiperspectivespatial}, leaf, text width=33em, edge=black, fill=red!20]]
                        [3D/Embodied Benchmarks , fill=red!20[ScanRefer\citep{chen2020scanrefer}{\,,} 
                       Multi3DRefer\citep{zhang2023multi3drefer}{\,,}
                        \\
                        RIORefer\citep{miyanishi_2024_3DV}{\,,}
                        GPT4Point Dataset\citep{qi2023gpt4point}{\,,}
                        \\
                        SQA3D\citep{ma2022sqa3d}{\,,}
                        ScanScribe\citep{zhu20233d}, leaf, text width=33em, edge=black, fill=red!20]]
                    ]
                    [
                        Evaluation Metrics \S ~\ref{em}, fill=red!20
                        [Traditional metrics, fill=red!20[Accuracy{\,,} F1 score{\,,} Recall{\,,} BLEU\citep{papineni2002bleu}{\,,}
                        \\
                        ROUGE\citep{lin2004rouge}{\,,} CIDEr\citep{vedantam2015cider}
                        EMD\citep{Erickson_2021}{\,,}
                        \\
                        CD\citep{wu2021densityawarechamferdistancecomprehensive}{\,,}
                        SPL\citep{yokoyama2023successweightedcompletiontime}, leaf, text width=33em, edge=black, fill=red!20]]
                        [Human and LLM evaluation, fill=red!20[GPTscore\citep{fu2023gptscoreevaluatedesire}{\,,}
                        LLM-Eval\citep{lin2023llmevalunifiedmultidimensionalautomatic}{\,,}
                        \\
                        AttrScore\citep{yue2023automaticevaluationattributionlarge}{\,,}
                        REVISEVAL\citep{zhang2025revisevalimprovingllmasajudgeresponseadapted}{\,,}
                        \\
                        MAJ-EVAL\citep{chen2025multiagentasjudgealigningllmagentbasedautomated}{\,,}
                        \cite{li2025largelanguagemodelsstruggle}{\,,}
                        \cite{liu2021doesmodelfailhumanintheloop}, leaf, text width=33em, edge=black, fill=red!20]]
                    ]
                ]
                [
                    Methods for \\improvement \S ~\ref{methods improve}, fill=green!20
                    [
                        Training-based methods \S ~\ref{tbm}, fill=green!20
                        [Spatial-Aware Module Training, fill=green!20[LLaVA-3D\citep{zhu2024llava}{\,,} 
                        Scene-LLM \citep{fu2024scene}{\,,}
                        \\
                        PointLLM\citep{guo2023point}{\,,}
                        SR-3D \citep{cheng20253dawareregionprompted}, leaf, text width=33em, edge=black, fill=green!20]]
                        [Synthetic Data for Task-Specific \\ Fine-Tuning, fill=green!20[SpatialVLM\citep{Chen_2024_CVPR}{\,,} 
                        SAT\citep{ray2025satdynamicspatialaptitude}, leaf, text width=33em, edge=black, fill=green!20]]
                        [Training Reasoning Processes with \\ Reinforcement Learning, fill=green!20[Pixel Reasoner\citep{su2025pixelreasonerincentivizingpixelspace}{\,,} 
                        Embodied-R \citep{zhao2025embodiedrcollaborativeframeworkactivating}{\,,}
                        \\
                       ManipLVM-R1\citep{song2025maniplvmr1reinforcementlearningreasoning}{\,,}
                        RoboRefer \citep{zhou2025roboreferspatialreferringreasoning}{\,,}
                        \\
                        SpaceR \citep{ogezi2025spareenhancingspatialreasoning}{\,,}
                        MetaSpatial \citep{pan2025metaspatialreinforcing3dspatial}, leaf, text width=33em, edge=black, fill=green!20]]
                    ]
                    [
                        Inference-based methods \S ~\ref{rbm}, fill=green!20
                        [Chain-of-thought Prompting and \\ Its Variants, fill=green!20[SpatialCoT\citep{liu2025spatialcotadvancingspatialreasoning}{\,,} 
                         spatialVLM\citep{Chen_2024_CVPR}{\,,}
                        \\
                        VoT\citep{wu2024minds}{\,,}
                        MVoT\citep{li2025imaginereasoningspacemultimodal}, leaf, text width=33em, edge=black, fill=green!20]]
                        [Explicit Spatial Representation, fill=green!20[SG-Nav\citep{yin2024sgnavonline3dscene}{\,,} 
                         Agent3D-Zero\citep{zhang2024agent3dzeroagentzeroshot3d}{\,,}
                        \\
                        \cite{wang2024dspybasedneuralsymbolicpipelineenhance}{\,,}
                        $SG^2$\citep{chen2025schemaguidedscenegraphreasoningbased}, leaf, text width=33em, edge=black, fill=green!20]]
                    ]
                ]
                [
                    Open Challenges \\and Future \\Directions \S ~\ref{open cad}, fill=orange!20
                    [
                        Challenges \S ~\ref{challenges}, fill=orange!20
                        [Deficiencies in Datasets and Benchmarks, text width=25em, fill=orange!20]
                        [Incomplete Spatial Understanding, text width=25em, fill=orange!20]
                        [Architectural and Training Paradigm Issues, text width=25em, fill=orange!20]
                    ]
                    [
                        Future Directions \S ~\ref{future directions}, fill=orange!20
                        [Building High Quality Datasets and Benchmarks, text width=25em, fill=orange!20]
                        [Developing Spatially-Aware Training Strategies, text width=25em, fill=orange!20]
                        [Exploring Novel Architectures for Spatial Intelligence, text width=25em, fill=orange!20]
                    ]
                ]
            ]
        \end{forest}
    }
    \caption{Taxonomy of our survey. We introduce a cognitive taxonomy of spatial reasoning tasks, organizing them by function and reasoning complexity. We also map existing benchmarks, review evaluation metrics, and analyze training- and reasoning-based methods to improve spatial ability. The study highlights key gaps and future directions toward developing models with more human-like spatial intelligence.}
    \label{fig:taxonomy}
\end{figure*}
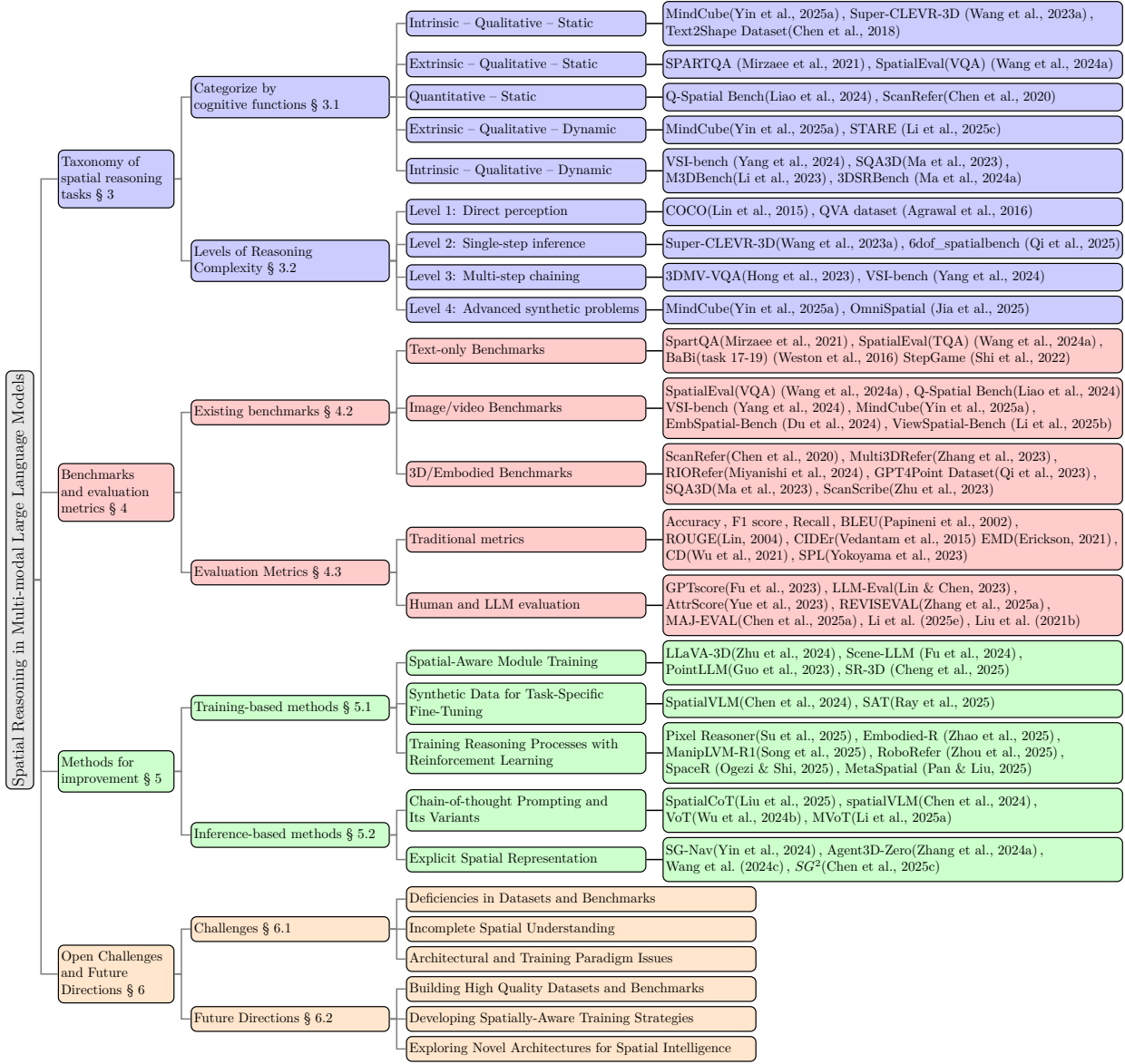

\noindent\textbf{   Comparison with existing survey:  } 
While recent comprehensive surveys have mapped the rapidly expanding landscape of 3D-capable LLMs, our work introduces a distinct organizational framework. For instance, \citet{2405.10255} provides a thorough meta-analysis that categorizes the field based on the various roles LLMs play in 3D tasks, such as scene understanding, captioning, and embodied navigation. More recently, \citet{2504.05786} is structured around the primary input modality, offering a taxonomy that distinguishes between image-based, point cloud-based, and hybrid approaches to grant LLMs 3D capacity.

In contrast, our survey departs from these modality-driven or task-centric taxonomies. We argue that true spatial intelligence ability is not solely determined by the input format or the specific application. Instead, it is determined by the underlying cognitive processes that govern how a model perceives, represents, and manipulates spatial information. These processes include how the model establishes frames of reference , the type of spatial information it uses, and whether it performs static understanding or dynamic mental transformation. Furthermore, spatial intelligence is also measured by the depth of reasoning complexity—whether a model can move beyond direct perception to perform multi-step inference, chaining, and compositional problem-solving. Based on this statement, we introduce a novel taxonomy organized from a cognitive function perspective. By analyzing spatial tasks in three fundamental dimensions and stratifying them across four levels of reasoning complexity. Our survey offers a more principled framework for analysis. This cognitive-centric approach enables deeper cross-task comparisons and is specifically designed to reveal critical gaps between current model capabilities and the nuances of human-like spatial intelligence.

\noindent\textbf{   Contributions:   }
In this paper, we provide a structured and insightful overview of spatial intelligence in MLLMs. As illustrated in Figure \ref{fig:taxonomy}, our main contributions are as follows: First, we introduce a novel taxonomy that organizes spatial tasks from a cognitive perspective, classifying them into five fundamental categories and four distinct levels of reasoning complexity. This framework moves beyond modality-based classifications to enable a more principled comparison of tasks and to reveal the critical gaps between current model capabilities and human-like spatial intelligence. Second, using this taxonomy as a foundation, we conduct a comprehensive survey of the current landscape, systematically mapping existing benchmarks across text-only, vision-language, and 3D settings to our proposed framework. We also review the diverse evaluation metrics and methodologies for robustly assessing spatial intelligence. Finally, we analyze and categorize methods for improving spatial intelligence into two main paradigms: training-based and inference-based approaches. This dual-perspective analysis clarifies their respective strengths and limitations, uncovering complementary mechanisms and cross-cutting trends. By synthesizing these elements, we aim to equip researchers with a thorough understanding of the field's current state and provide intuition on possible directions for future works.

\section{Background Knowledge and motivation}

\subsection{Modern Transformer-based Models}
The architectural bedrock for modern large-scale models is the Transformer\citep{10.5555/3295222.3295349}. Its core innovation is the attention mechanism, which processes all input tokens in parallel, unlike the sequential processing of its predecessors like RNNs\citep{ELMAN1990179}. This mechanism enables the model to adaptively assign contextual relevance to all tokens within a sequence when encoding a particular token, thereby facilitating the efficient modeling of complex and long-range dependencies. To preserve sequence information, the architecture incorporates positional encodings. This inherently parallelizable architecture, coupled with its strong capacity for contextual modeling, provides MLLMs with strong ability to perceive various information and generate reasonable results.

\subsubsection{Large Language Models} 
Large Language Models (LLMs) represent a paradigm shift in artificial intelligence. They are built upon Transformer architecture and scaled to unprecedented sizes, often containing hundreds of billions of parameters. Trained on extensive web-scale corpora of text and code, these models learn to predict the next token in a sequence, a seemingly simple objective that yields a remarkably sophisticated understanding of linguistic structure and meaning. Prominent examples such as the GPT series, LLaMA\citep{touvron2023llama}, and PaLM\citep{chowdhery2023palm} have demonstrated remarkable emergent abilities that go far beyond simple text generation. These abilities include few-shot and zero-shot in-context learning, where they can perform new tasks with only a handful of examples, and complex reasoning through techniques like chain-of-thought prompting.

Despite their strong linguistic capabilities, LLMs remain fundamentally limited in spatial reasoning, as their learning is confined to text-based representations derived from language-only training. Lacking any sensory perception, LLMs learn spatial concepts as statistical patterns of words, not as grounded geometric representations. While this allows them to handle simple categorical relations (e.g., on, next to), their understanding often lacks metric precision and physical consistency. A core focus of current research is to bridge this gap between language-based reasoning and a robust, grounded comprehension of the spatial world.

\subsubsection{Vision-Language Models}
Vision-Language Models (VLMs) extend the capabilities of LLMs by integrating visual data, enabling them to reason jointly across both images and text. From an architectural perspective, these models commonly integrate a pre-trained vision encoder, such as the Vision Transformer (ViT)~\citep{dosovitskiy2021an}, which converts visual inputs into sequences of embeddings, with a large language model for visual reasoning. A specialized alignment module and cross-attention mechanism is used to bridge these two modalities, creating a shared representation space where visual concepts are mapped to linguistic ones. Pioneering models like CLIP\citep{radford2021learningtransferablevisualmodels} demonstrated the power of this alignment through contrastive learning, while more recent architectures like LLaVA\citep{liu2023visualinstructiontuning} and Flamingo\citep{alayrac2022flamingovisuallanguagemodel} have enabled sophisticated multi-modal dialogue and instruction-following.

From a spatial intelligence perspective, VLMs represent a significant step forward from their text-only counterparts. They can ground spatial language (e.g ``the blue sphere to the left of the green cube'') in the actual pixel space of an image, associating words with specific visual regions. This allows them to reason about 2D relationships like relative position, alignment, and basic object interactions within a single view. However, the understanding of most VLMs is confined to this 2D projective plane. They inherently struggle to infer 3D spatial properties such as depth, volume, and the relationships between occluded objects from a single image. Thus, while VLMs provide a crucial visual anchor for language, a central research frontier lies in extending this 2D-grounded understanding to a comprehensive and robust 3D spatial intelligence.

\subsection{Cognitive Functions}
To systematically analyze the spatial capabilities of large language models, it is essential to first ground our discussion in the principles established by cognitive science. Human spatial cognition is not a monolithic process but a complex interplay of different representational systems and reasoning abilities. By deconstructing spatial abilities into core dimensions, we can create a more precise framework for evaluating AI models, identifying their specific strengths and weaknesses, and charting a path toward more human-like intelligence\citep{cogni}.

\begin{figure}[h]   
  \centering
  \includegraphics[width=1\linewidth]{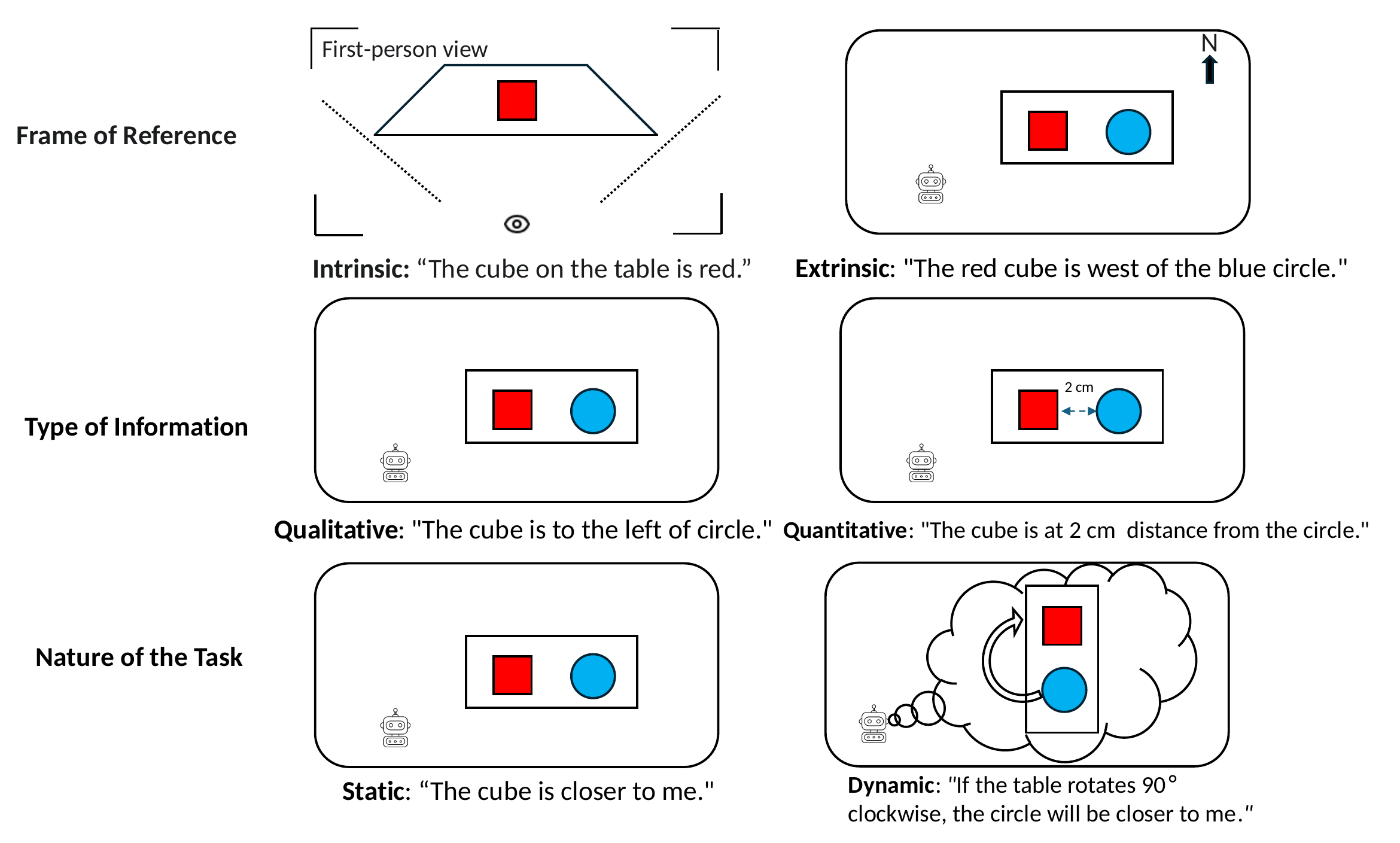}
  \caption{Illustration of cognitive dimensions: Spatial reasoning can be decomposed along three cognitive dimensions: frame of reference (intrinsic vs. extrinsic), type of information (qualitative vs. quantitative), and nature of the task (static vs. dynamic). Each dimension reflects a distinct way humans and models encode, compare, or transform spatial relations.}
  \label{fig:cognitive}
\end{figure}

As illustrated in Figure \ref{fig:cognitive}, this section introduces three fundamental, orthogonal dimensions that form the basis of our cognitive taxonomy: (1) the Frame of Reference used to anchor spatial relations, (2) the Type of Information being processed, and (3) the Nature of the Task being performed.

    \noindent\textbf{  Intrinsic vs. Extrinsic: }
A frame of reference is the coordinate system used to define and interpret the position, orientation, and relationship of objects. The choice of frame is critical, as it determines how spatial information is encoded and communicated.\citep{egocen, Klatzky1998} An intrinsic frame describes an object based on the inherent properties, orientation or parts of it. It mainly focuses on the object itself. While extrinsic frame also focuses on other objects in the scene or the properties within the scene.

    \noindent\textbf{  Qualitative vs. Quantitative: }
Spatial relationships can be described with varying levels of precision, ranging from qualitative abstractions to exact quantitative measurements. Quantitative reasoning involves processing spatial information that is continuous, and precise. While qualitative reasoning involves processing spatial information that is discrete, and abstract.\citep{DBLP:reference/fai/CohnR08, quanti} It simplifies the world into a set of relational categories, such as topology (on, inside, under), relative position (left of, next to, between), and orientation (parallel to, facing). A statement like ``the keys are on the table next to the book'' is a prime example of qualitative reasoning.

\noindent\textbf{  Static vs. Dynamic: }
The final dimension distinguishes between tasks that involve understanding a fixed scene and those that require mentally manipulating it.\citep{HEGARTY2004280,Newcombe2015ThinkingAS}Static reasoning concerns the description and comprehension of spatial relationships within a single, unchanging scene or configuration. It is foundational and serves as a prerequisite for more complex spatial cognition. 

Dynamic reasoning is a more advanced form of reasoning that involves mentally simulating changes in spatial relationships. It requires manipulating objects, viewpoints, or configurations in one's mind. It is the cornerstone of planning, problem-solving, and counterfactual thinking about the physical world. For AI models, this represents a significant leap from passive description to active, predictive simulation of spatial dynamics.

\subsection{From Perception to Reasoning}
Human spatial intelligence develops through a progressive hierarchy that transforms raw sensory input into abstract, manipulable mental representations.\citep{marr2010vision, 10.5555/7909} Three interdependent words are always mentioned: spatial perception, spatial understanding, and spatial reasoning. Spatial perception is the lowest-level process, responsible for acquiring and organizing raw data from the environment. This involves the neuro-biological processing of inputs from vision, touch, and hearing. Building upon perception, spatial understanding integrates discrete sensory impressions into a coherent internal representation of the environment, forming a mental map or cognitive schema that captures the relational structure among objects.\citep{tolman1948cognitive, eichenbaum2015hippocampus} Spatial reasoning involves the active manipulation of internal spatial representations to imagine transformations, predict outcomes, and solve problems. This is a dynamic process that operates on the internal model created through understanding. It requires the ability to simulate motion, rotation, or perspective change within one’s mental model.\citep{Newcombe2015ThinkingAS}

Together, these three layers form a continuum of increasing cognitive complexity. Progress along this continuum reflects a shift from direct sensory encoding toward abstract, generative manipulation of spatial knowledge. In LLMs, advancing from perceptual recognition to high-level reasoning requires similar transitions: from extracting explicit spatial cues, to constructing stable internal representations, and ultimately to operating on those representations to achieve flexible, predictive, and physically consistent understanding of space.

\subsection{Challenges of MLLM for spatial reasoning}
The extension from language-only models to MLLMs aims to endow systems with grounded perception and spatial awareness. By coupling visual encoders with linguistic reasoning modules, MLLMs promise to bridge the gap between language-based reasoning and the geometric regularities of the physical world. Despite this integration, current vision-language and multimodal architectures still fall short of achieving robust spatial intelligence. The limitations arise not only from the textual bias inherited from large language models, but also from several critical failure modes shown below:

\noindent\textbf{The Projection Bottleneck of Visual Encoders:}
Modern VLMs begin with 2D encoders that tokenize images into patches optimized primarily for semantic alignment with language rather than faithful 3D geometry. After cross-modal projection, these visual tokens enter the language space as a flattened sequence, where 3D information like fine-grained depth ordering, orientation, and metric continuity are weakly preserved. Empirically, diagnostic studies show that models can correctly name objects yet fail on queries requiring precise spatial structure, indicating that spatial cues are not robustly carried\citep{wang2024spatial}. Multi-view settings further expose this bottleneck: features aligned per image seldom aggregate into a consistent scene-level representation, leading to contradictions across views\citep{li2025viewspatialbenchevaluatingmultiperspectivespatial}.

\noindent\textbf{Learning Statistical Correlations Instead of Physical Constraints:}
Pretraining objectives bias MLLMs to exploit semantic co-occurrence (``cup on table'') rather than obeying geometric or physical regularities. As a result, models often answer spatial questions by pattern completion instead of checking consistency with the visible scene or with simple physics. Benchmark evidence highlights this shortcutting behavior. Model performance drops sharply on metric or counterfactual questions, and even chain-of-thought can amplify problematic steps in spatial tasks \citep{yang2024think}. Quantitative probes focused on metric reasoning report systematic brittleness compared to counting\citep{liao2024reasoningpathsreferenceobjects}, and dynamic or 6-DoF evaluations reveal violations of plausible motion or contact constraints. Attention analyses further show that models prioritize salient semantics over geometry-bearing regions, explaining confident yet spatially incorrect outputs\citep{chen2025spatialreasoninghardvlms}.

\noindent\textbf{Ambiguity and Frame-of-Reference Instability:}
Spatial information is viewpoint-relative (egocentric) or world-relative (allocentric), and many tasks require switching or maintaining consistency across multiple camera poses. Current MLLMs have no explicit mechanism for reference-frame management. Instead, they rely on attention over mixed tokens, which leads to drift when the viewpoint changes. When tests involve multi-view localization or perspective-taking, models often flip left and right or front and behind. Object localization task achieves less accuracy when the camera moves\citep{li2025unfoldingspatialcognitionevaluating}. A model may look strong on a single image, but when multiple views are combined its references become unstable and it lacks persistent scene memory, leading to conflicting answers about the same scene from different viewpoints.

The challenges faced by MLLMs extend beyond the language domain. They originate from a fundamental representational gap between linguistic abstraction and geometric continuity, compounded by architectural constraints that compress high-dimensional perception into discrete tokens. Overcoming these deficiencies requires models that integrate continuous spatial structures, persistent memory, and physically grounded reasoning, as well as bridging perception and cognition in a way that mirrors human spatial understanding.

\section{Taxonomy of Spatial Reasoning Tasks}
\label{taxonomy label}
To provide a systematic framework for analyzing the diverse landscape of spatial reasoning tasks, this section introduces a novel taxonomy designed to move beyond classifications based purely on input modality. Current benchmarks are often grouped by whether they are text, image or 3D input, a categorization that fails to capture the core cognitive challenge a task presents. Instead, our taxonomy is built upon two more fundamental, orthogonal dimensions: a classification of tasks into five broad categories based on their underlying cognitive function, and a hierarchy of these tasks across four levels of reasoning complexity. This dual-axis framework allows for a more principled comparison of different benchmarks, helps to diagnose model capabilities more precisely, and systematically reveals the key research challenges that lie ahead.

\subsection{A Cognitive Taxonomy of Spatial Tasks}
\label{cognitive function categorization}
By combining the three cognitive dimensions: Frame of reference (intrinsic vs. extrinsic), type of information (quantitative vs. quantitative) and nature of the task (static vs. dynamic), we can derive a comprehensive taxonomy. We focus on the five categories that represent the most distinct and challenging axes of spatial cognition found in modern benchmarks. Three categories are excluded because they either don't have meaningful aspects or are too similar to categories selected. This set prioritizes unique cognitive challenges, such as the distinction between static scene description and dynamic mental transformation, providing a clear and powerful framework for analyzing the capabilities of LLMs and VLMs.

\noindent\textbf{Intrinsic – Qualitative – Static:}
This foundational category involves reasoning about the internal structure and property of a single, unchanging object. The frame of reference is the object itself. It tests a model's ability to understand spatial knowledge within the object. ``The chair's back is above its seat'' is an easy example in this category.

\noindent\textbf{Extrinsic – Qualitative – Static:}
This is the most common category in spatial reasoning research. Lots of benchmarks have corresponding samples. It focuses on describing the arrangement of objects within a static scene using qualitative, relational terms. The frame of reference is external. As a result, the question will involve multiple objects or elements in the scene. Question like ``What is positional relationship between the table and the lamp?'' falls in this category.

\noindent\textbf{Quantitative – Static:}
This category consolidates the challenge of metric reasoning. We don't specify the first dimension here because the quantitative reasoning ability is the same no matter for intrinsic or extrinsic cases. It requires a model to process precise, quantitative information about a static scene, applying to both the intrinsic properties of a single object and the extrinsic relationships between multiple objects. The core challenge is the model's ability to handle continuous spatial data. One typical example is ``What is the height of this table in meters?''

\noindent\textbf{Intrinsic – Qualitative – Dynamic:}
This advanced category involves the mental transformation of an object's parts based on its structure. It requires simulating how an object's configuration changes through manipulation, which is a challenging process, especially for LLMs with limited training data in physically operating objects. \citet{li2025unfoldingspatialcognitionevaluating} gives question about folding a cube, which requires ability in this category to solve.

\noindent\textbf{Extrinsic – Qualitative – Dynamic:}
This final category also involves dynamic reasoning, but the transformation concerns the object's relationship with other object or the whole environment. It requires mentally simulating a change in viewpoint or position within a larger scene. Perspective changing question like ``If I'm sitting on the sofa, what's the object on my right?'' falls in this category. This is also a challenging category for current LLMs.

By organizing tasks along these cognitive dimensions, we create a more insightful and robust analytical framework.For instance, many benchmarks contain image QA tasks. But these questions varies in cognitive aspect. They can be static description or profoundly mental transformation, which is well classified in our taxonomy. This allows for a more principled comparison across benchmarks, a more precise diagnosis of model failures, and a clearer identification of which cognitive skills are under-represented in current research. In essence, this framework shifts the focus from what a model perceives to how it reasons, offering a more stable and forward-looking foundation for advancing true spatial intelligence.

\subsection{Levels of Reasoning Complexity}
\label{reasoning complexity level}
This dimension of our taxonomy describes the complexity of the cognitive process required to solve a task. We define four distinct levels, moving from simple information retrieval to complex, multi-step problem-solving. This hierarchy allows us to measure the ``cognitive depth'' of a task, providing a more comprehensive understanding of a model's capabilities and limitations.

\noindent\textbf{Level 1: Direct Perception:}
This is the most fundamental level, involving the direct retrieval and description of explicit information from the input. It requires no inference or manipulation of spatial relationships; the answer can be ``read directly'' from the scene representation. This level tests the model's core perceptual abilities, such as object recognition, attribute identification, and scene awareness.

\noindent\textbf{Level 2: Single-Step Inference:}
This level introduces a single, simple layer of reasoning. It requires the model to go beyond direct perception to infer a basic spatial relationship between two or more objects or entities. While the objects themselves are directly perceived, their relationship is not explicitly stated and must be deduced.

\noindent\textbf{Level 3: Multi-Step Chaining:}
At this level, a task cannot be solved with a single inference. It requires a sequence of chained reasoning steps, where the conclusion of one step serves as a premise for the next. This is analogous to ``chain-of-thought'' reasoning but applied to a spatial context. It often involves decomposing a complex query into a series of simpler sub-problems and maintaining a mental state throughout the process.

\noindent\textbf{Level 4: Advanced Synthetic Problems:}
This highest level of complexity involves tasks that synthesize general reasoning parts, designed specifically to test the limits of a model's reasoning and generalization abilities. These problems typically require a combination of different reasoning types (e.g.integrating spatial reasoning with common sense reasoning) and often involve complex dynamic transformations that cannot be solved by simply applying learned knowledge. For example, ``If the stack of blocks is pushed from the right, in what order will they fall?'' This level serves as a representation for measuring more general intelligence in the spatial domain.

This stratification by reasoning complexity is essential because it provides a measure of a task's cognitive depth, moving beyond a simple pass or fail evaluation. It allows for a more granular diagnosis of a model's capabilities: a model that excels at single-step inference but consistently fails at multi-step chaining has a specific, identifiable deficit in its planning and sequential reasoning abilities, not just a general weakness in spatial understanding. When combined with our five cognitive categories, this hierarchy creates a comprehensive two-dimensional grid for classifying any spatial task. This framework enables a more rigorous and meaningful assessment of progress in the field, allowing us to track how models are advancing from basic perception towards more sophisticated problem-solving abilities.

\subsection{Illustrative Examples }
\begin{figure}[h]   
  \centering
  \includegraphics[width=1\linewidth]{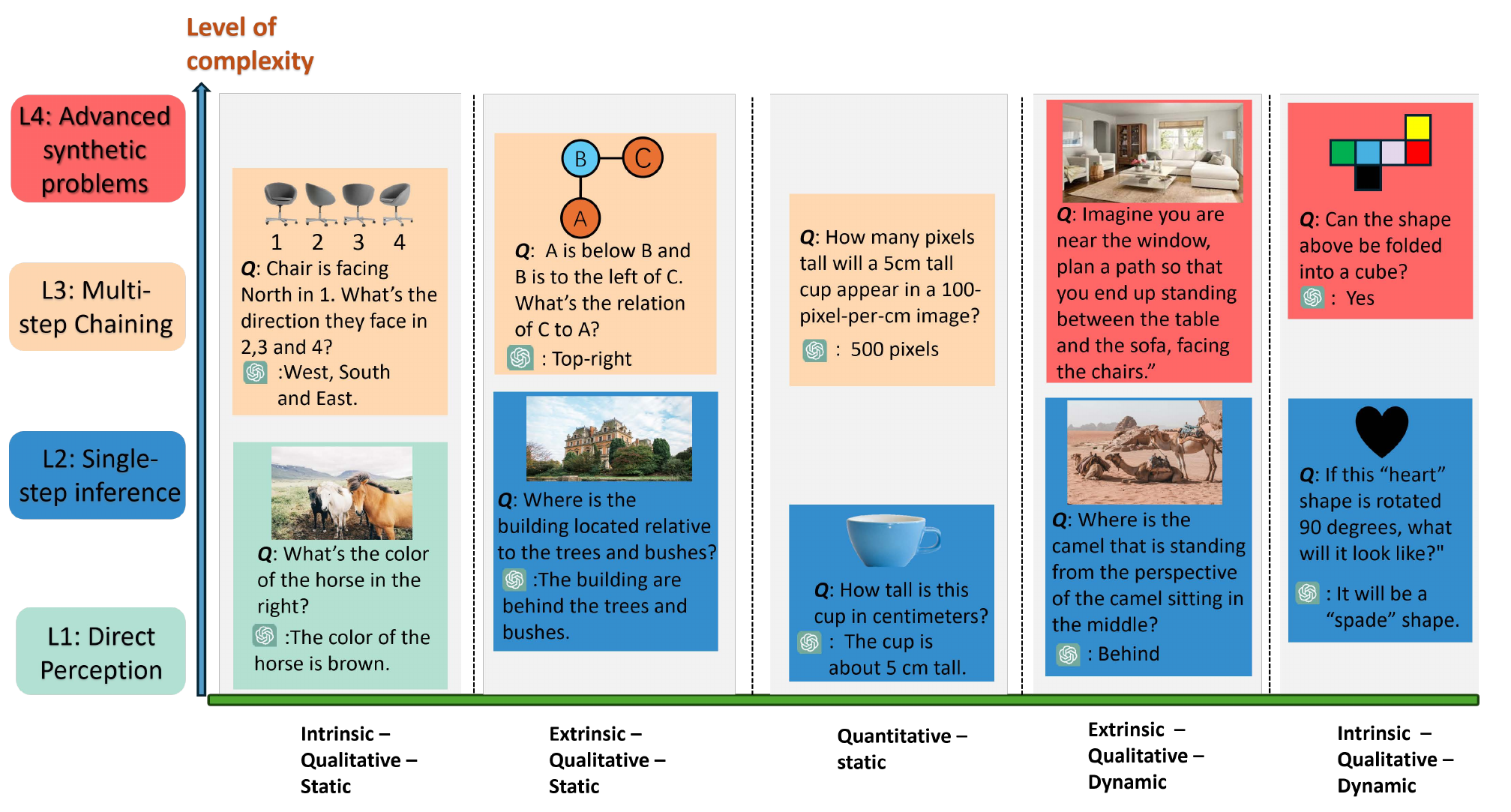}
  \caption{Illustrative Examples for the Cognitive and Complexity-Based Taxonomy: This figure maps representative spatial reasoning tasks across five cognitive categories (x-axis) and four levels of reasoning complexity (y-axis). The taxonomy progresses from direct perception to advanced synthetic reasoning, distinguishing intrinsic vs. extrinsic, static vs. dynamic, and qualitative vs. quantitative cognition. Together, it illustrates how task complexity and cognitive function jointly define the difficulty and nature of spatial reasoning challenges for MLLMs.}
  \label{fig:illutrateto}
\end{figure}

To ground the concepts of our two-dimensional taxonomy, Figure \ref{fig:illutrateto} provides a set of illustrative examples, mapping various spatial tasks onto the grid created by the five cognitive categories and the four levels of reasoning complexity. The horizontal axis represents the five fundamental cognitive categories, which define the type of spatial knowledge being processed. The vertical axis stratifies these tasks by their reasoning complexity, defining the cognitive process required for their solution. This dual-axis approach provides a nuanced framework for analyzing model capabilities.

The power of this framework is evident when observing how different tasks are classified. For instance, consider the vertical progression within a single category. In the Extrinsic-Qualitative-Static column, a Level 2 task requires a simple deduction, such as determining that ``the building is behind the trees.'' In contrast, the Level 3 task involving objects A, B, and C requires a sequence of deductions: the model must first process ``A is below B'', then integrate ``B is to the left of C'', and finally synthesize these two premises to deduce the composite relationship between C and A, which is top-right.

Similarly, the framework reveals how different cognitive skills can exist at the same level of complexity. At Level 2, for example, the task of locating the camel in Extrinsic-Qualitative-Dynamic and the task of mentally rotating the heart shape in Intrinsic-Qualitative-Dynamic both require a single inferential step. However, they test fundamentally different abilities. One requires consideration with other object, the other requires imagination of manipulation within one object.  A model might easily succeed at one while failing at the other, although they are at different complexity levels.

This figure also clarifies our definition for different cognitive categories. For example, the chair orientation problem is classified as Level 3 and static because all potential outcomes are visually present. The challenge is not in imagining the rotation from scratch, but in applying a rule ``Chair 1 is North'' through a multi-step deductive process to correctly label the existing images. And for level 4 questions, as illustrated by the path planning and cube folding examples, this level denotes problems where the solution cannot be derived from perception alone. These tasks demand the application of abstract rules like the geometry of cube folding, complex planning under constraints like pathfinding, or integrated commonsense knowledge, representing a significant leap toward more general, fluid intelligence.

In conclusion, this two-dimensional categorization provides a far more granular and insightful diagnostic tool than a classification based on input modality. It allows us to distinguish between a model's ability to perceive a scene and its ability to reason about it. By pinpointing whether a model’s weaknesses lie in specific cognitive categories or in managing higher levels of reasoning complexity, this framework offers a clear and actionable path for the targeted development of spatially intelligent AI.

\section{Datasets, Benchmarks and Evaluation metrics}
\label{b and em}
\subsection{Benchmarks are Tools that Shape the Field}
\begin{table*}[t]
\centering
\begin{adjustbox}{max width=\textwidth}
\begin{tabular}{lcc|ccccc|c}
\toprule
 & \multirow{2}{*}{\textbf{\Large Benchmark}}
 & \multirow{2}{*}{\textbf{\Large Main Tasks}}
 & \multicolumn{5}{c}{\textbf{\Large Cognitive functions}} 
 & \multirow{2}{*}{\textbf{\Large Environment}} \\
\cmidrule(lr){4-8}
 &  &  
 & \textbf{\large I-Ql-S} 
 & \textbf{\large E-Ql-S} 
 & \textbf{\large Qn-S} 
 & \textbf{\large I-Ql-D} 
 & \textbf{\large E-Ql-D} 
 &  \\
\midrule
 
\multirow{4}{*}{\rotatebox{90}{Text}}    
& SPARTQA \citep{mirzaee-etal-2021-spartqa} & Attribute,relational QA & -  & L3 & L1 & - & - & Synthetic \\
& SpatialEval(TQA) \citep{wang2024spatial} & Counting,relational QA & -  & L3 & L1 & - & L4 & Synthetic \\
& BaBi(task 17-19) \citep{DBLP:journals/corr/WestonBCM15} & Relational QA, Navigation & L2  & L2 & L3 & - & L4 & Synthetic \\
& StepGame \citep{stepGame2022shi} & Relational QA & -  & L3 & - & - & L2 & Synthetic \\

\midrule
\multirow{20}{*}{\rotatebox{90}{Image or video}}  
& Super-CLEVR-3D \citep{wang20233d} & Attribute, relational QA & L2  & L2 & -  & - & - & Synthetic \\

& Open3DVQA \citep{zhang2025open3dvqa} & Attribute, relational QA & L2  & L2 & L3  & - & L3 & Synthetic \\ 
& LAMM(Image)\citep{yin2023lamm} & Captioning, relational QA & L2  & L3 & L2 & - & L1 & Real-world, Synthetic \\
& 3DMV-VQA\citep{hong20233d} & Counting, relational QA & L2  & L3 & L1 & - & L3 & Real-world \\
& MindCube\citep{yin2025spatialmentalmodelinglimited} & Relational, dynamic QA & L4  & L3 & - & L4 & L3 & Real-world \\
& NuScenes-QA\citep{qian2023nuscenes} & Counting, relational QA & L2  & L2 & L1 & - & L3 & Real-world \\
& Q-Spatial Bench\citep{liao2024reasoningpathsreferenceobjects} & Metrics QA & -  & - & L3 & - & - & Real-world \\
& SpatialEval(VQA) \citep{wang2024spatial} & Relational QA, Navigation & -  & L3 & L1 & - & L4 & Real-world, Synthetic \\
& EmbSpatial-Bench \citep{du-etal-2024-embspatial} & Relational QA & -  & L2 & - & - & L3 & Real-world, Synthetic \\
& 6dof\_spatialbench \citep{sofar25} & Counting, relational QA & L2  & L2 & L1 & - & - & Real-world \\
& 3DSRBench \citep{ma20243dsrbench} & Relational, orientation QA,  & -  & L2 & L3 & - & L4 & Real-world \\
& STAR \citep{wu2021star_situated_reasoning} & Relational, dynamic QA & -  & L2 & - & - & L4 & Real-world \\
& STARE \citep{li2025unfoldingspatialcognitionevaluating} & Dynamic QA & L1  & L1 & L1 & L4 & L4 & Real-world, Synthetic \\
& SAT \citep{ray2025satdynamicspatialaptitude} & Counting, relational QA & -  & L2 & L1 & - & L4 & Real-world, Synthetic \\
& Spatial-MLLM-120k \citep{wu2025spatial} & Counting, relational QA & L1  & L2 & L1 & - & L3 & Real-world \\
& SPHERE \citep{zhang2024sphere} & Relational QA, Localization & -  & L2 & L2 & - & L4 & Real-world \\
& Spatial457 \citep{wang2025spatial457diagnosticbenchmark6d} & Counting, relational QA & L2  & L3 & L1 & - & L2 & Synthetic \\
& ViewSpatial-Bench \citep{li2025viewspatialbenchevaluatingmultiperspectivespatial} & Dynamic QA, Localization & -  & L2 & - & - & L4 & Real-world \\
& OmniSpatial \citep{omnispatial25} & Relational QA, Navigation & L2  & L2 & L2 & L4 & L4 & Real-world, Synthetic \\
& VSI-bench \citep{yang2024think} & Relational QA, Navigation & L2  & L3 & L3 & - & L4 & Real-world \\
& Ego3D-Bench \citep{gholami2025spatialreasoningvisionlanguagemodels} & Relational QA, Metrics QA & L2  & L3 & L4 & - & L3 & Real-world \\

\midrule
\multirow{13}{*}{\rotatebox{90}{3D files}} 
& Multi3DRefer\citep{zhang2023multi3drefer} & 3D grounding & L2  & L3 & - & - & - & Real-world \\
& ScanRefer\citep{chen2020scanrefer} & 3D grounding & L2  & L3 & L3 & - & L3 & Real-world \\
& Chat-3D Dataset\citep{wang2023chat3d} & 3D captioning, Conversation & L2  & L3 & L2 & L3 & L4 & Real-world \\
& Text2Shape Dataset\citep{chen2018text2shape} & 3D generation & L2  & - & L1 & - & - & Synthetic \\
& Cap3D Dataset\citep{luo2023scalable} & 3D captioning & L2  & L2 & - & - & - & Synthetic \\
& SceneVerse\citep{jia2024sceneverse} & 3D grounding, Relational QA & L2  & L3 & L1 & - & - & Real-world, Synthetic \\
& RIORefer\citep{miyanishi_2024_3DV} & 3D grounding & L2  & L3 & - & - & L3 & Real-world \\
& M3DBench\citep{li2023m3dbench} & Captioning, Navigation & L2  & L3 & L2 & L3 & L4 & Real-world \\
& GPT4Point Dataset\citep{qi2023gpt4point} & Captioning, 3D generation & L2  & L2 & L2 & - & - & Synthetic \\
& LAMM(3D)\citep{yin2023lamm} & Object detection, Relational QA & L2  & L3 & L2 & - & - & Real-world, Synthetic \\
& ScanScribe\citep{zhu20233d} & Grounding, Captioning, QA & L2  & L3 & L2 & L3 & - & Real-world \\
& CLEVR3D\citep{yan2023comprehensive} & Counting, relational QA & L1  & L2 & L1 & - & - & Real-world, Synthetic \\
& SQA3D\citep{ma2022sqa3d} & Counting QA, Localization & -  & L2 & L1 & - & L4 & Real-world \\

\bottomrule
\end{tabular}
\end{adjustbox}
\caption{Table of benchmarks for LLM spatial reasoning. This table provides a comprehensive overview of existing benchmarks, including primary modality, main tasks, cognitive functions, and task complexity levels. We also include details on environment (whether the data is real-world or synthetically generated). The cognitive categorization follows the cognitive taxonomy and level of complexity introduced in section 3, where I-Ql-S means Intrinsic – Qualitative – Static. E-Ql-S means Extrinsic – Qualitative – Static. Qn-S means Quantitative – static. I-Ql-D means Intrinsic – Qualitative – Dynamic. E-Ql-D means Extrinsic – Qualitative – Dynamic. L1,L2,L3,L4 correspond to four levels of complexity.}
\label{tab:benchmarks}
\end{table*}

The evaluation of spatial reasoning in Large Language Models is fundamentally reliant on a sophisticated ecosystem of datasets and benchmarks. These tools are not merely for ranking models; they provide the structured environments necessary to dissect the multifaceted nature of spatial intelligence—from basic perception to complex, dynamic reasoning. At their foundation are datasets, the curated collections of annotated visual or 3D data that serve as the ground truth of the physical world. Built upon these are benchmarks, which define the specific challenges and evaluation protocols used to probe a model's capabilities. For LLMs, which learn spatial concepts from abstract textual patterns, these benchmarks are the primary mechanism for testing the validity, consistency, and geometric grounding of their knowledge. They force a confrontation between a model's linguistic plausibility and the physical realities of a scene, making them indispensable for measuring genuine progress.

The current landscape of these benchmarks is rapidly expanding, reflecting the community's growing ambition to move beyond simple 2D understanding. To illustrate this diverse body of work, this section offers a comprehensive analysis guided by the cognitive taxonomy introduced in Section 3. The analysis is supported by a detailed overview in Table \ref{tab:benchmarks}, which characterizes prominent benchmarks according to their data sources, primary tasks, and, most importantly, a cognitive profile derived from our cognitive categories and levels of complexity. This structured approach allows for an examination of overarching trends in the field, revealing the dominant research paradigms and identifying critical gaps in how we currently challenge and measure the spatial intelligence of AI.
\subsection{The Landscape of Spatial Reasoning Dataset and Benchmarks}
\label{eb}
A high-level analysis of the benchmark landscape, as illustrated in Table \ref{tab:benchmarks}, reveals distinct patterns in how the research community has approached the evaluation of spatial reasoning in LLMs. Rather than a uniform distribution of challenges, the field has concentrated its efforts in specific areas while leaving some others underexplored. We will analyze these trends, using our cognitive taxonomy to explain the current state of the field and to identify its most critical frontiers.

\subsubsection{Prevalence of relational reasoning questions}
The most prominent trend is the significant concentration of benchmarks within the \textbf{Extrinsic-Qualitative-Static} category, which are mostly relational reasoning questions. For example, \cite{mirzaee-etal-2021-spartqa}  designed geometric rules to automatically generate a text description of visual scenes and corresponding spatial QA pairs. \cite{wang20233d} explore 4 different factors in VQA domain isolated in order that their effects to relational reasoning problems can be studied. \cite{zhang2023multi3drefer} created a dataset that generalizing from ScanRefer\citep{chen2020scanrefer} to grounding multiple objects that described by their physical attributes or positions. The primary challenge in these benchmarks is to correctly associate natural language descriptions of positions with specific objects or regions based on their spatial relationships—such as what is ``on'', ``next to'', or ``behind'' something else. 

This focus is a natural consequence of the strengths of current models. Language is inherently well-suited to expressing qualitative relations, and vision-language models have become accustomed to grounding these descriptions in a visual way. These tasks represent a foundational and necessary step toward spatial intelligence, testing a model's ability to build a basic, semantically rich understanding of a static scene. 

\subsubsection{Unbalanced distribution in quantitative tasks}
While at first glance the Quantitative-Static category appears well-represented across many benchmarks, a closer analysis reveals that it is almost exclusively addressed at a superficial level. The dominant form of quantitative task is object counting, a common feature in many VQA datasets. This task, while valuable, can only be considered as Level 1 (Direct Perception), as it requires enumerating directly perceived instances rather than reasoning about metric properties. This prevalence of low-level counting tasks masks a deeper and more critical gap: the scarcity of benchmarks that test true metric reasoning—the ability to estimate, compare, or calculate continuous spatial properties like distance, size, angle, or volume. One good example is Q-Spatial Bench\citep{liao2024reasoningpathsreferenceobjects}. It explore quantitative spatial reasoning by questions asking about width, height or distance between objects.

The gap of lacking high level quantitative tasks exposes a core architectural and conceptual challenge: the fundamental mismatch between the discrete, token-based nature of LLMs and the continuous, metric nature of physical space. It is profoundly more challenging to train a model to map ambiguous linguistic phrases to precise geometric quantities than it is to count objects. Consequently, the field has a significant quantitative blind spot, directly bottlenecking the development of LLMs for applications like robotics, AR/VR, and autonomous navigation, where a precise physical understanding is not just beneficial but mission-critical.

\subsubsection{The Frontier of Dynamic Reasoning}
The most challenging evaluations of spatial intelligence are found in benchmarks targeting the dynamic and transformational categories. These tasks move beyond static description to require mental simulation, a key marker of deeper cognitive processing. This frontier is developing along two distinct axes. First, Intrinsic-Qualitative-Dynamic benchmarks, like \cite{li2025unfoldingspatialcognitionevaluating}, use problems like cube net folding to isolate the cognitive skill of mental simulation. Second, Extrinsic-Qualitative-Dynamic benchmarks evaluate a model's ability to reason about its relation to an environment. For example, \cite{ma2022sqa3d} created situated VQA tasks in 3D point cloud centered on perspective changing problems. Moreover, \cite{omnispatial25} built various tasks on navigation, object rotating and geometric reasoning. These challenges are essential for any embodied agent, as they require a model to dynamically update its internal world model from a new viewpoint. Success on these dynamic benchmarks, which often target the highest levels of reasoning complexity (L3 and L4), is a much stronger indicator of generalizable spatial intelligence than performance on static description tasks alone.

\subsubsection{Trends in Data}
\noindent\textbf{benchmark modality:}
While the cognitive dimension of a benchmark is defined by the reasoning skills it targets, the choice of data modality influences how these skills are elicited and evaluated. For instance, 3D files make it easier to design tasks that involve metric precision, whereas image-based QA emphasizes perceptual questions under partial observability like perspective changing problems. Text-only benchmarks often abstract away perception entirely, isolating reasoning patterns in linguistic form. Thus, modality should be seen not as determining the nature of reasoning but as a way through which reasoning tasks are created, defining task difficulty, ambiguity, and robustness.

\noindent\textbf{Synthetic and real-world datasets:}
The choice between synthetic and real-world environments reveals a trade-off between efficiency and effectiveness. Synthetic datasets like Open3DVQA\citep{zhang2025open3dvqa} offer perfect, noise-free ground truth and programmatic control, making them ideal for isolating specific reasoning skills. A wide range of tools and frameworks have been developed to create synthetic datasets for spatial reasoning, offering programmatic control over scenes, objects, and rendering. Classical engines such as \textbf{Unity} and \textbf{Unreal Engine} are widely used for building interactive 3D environments and simulating embodied tasks. More recent frameworks, such as ProcTHOR\citep{deitke2022procthorlargescaleembodiedai} and Infinigen \citep{infinigen2023infinite}, enable procedural generation of diverse indoor and outdoor scenes, with automatic annotations for geometry, segmentation, and camera viewpoints.

\cite{kabir2024comprehensivesurveyvisualquestion} mention that Synthetic datasets are easier, less expensive, and less time-consuming to produce as the building of a large dataset can be automated. And they can be tailored so that performing well on them requires better reasoning and composition skills. But training on synthetic dataset might bring about the problem of over-fitting or model collapse\citep{dohmatob2024strong}. Because they always contain some subtle patterns hidden in the synthetic process, causing the model to rely on artificial clues. In contrast, datasets built on real-world are more effective in training models for real-world applications. But they are prone to introduce perceptual challenges like noise, which requires more human effort to configure the scene and view. In conclusion, while they are more efficient and controllable, synthetic datasets are not sufficient alone for building robust VQA models and are best viewed as complementary to large-scale real datasets.

\subsection{Evaluation metrics}
\label{em}
Evaluation metrics play a central role in shaping how progress in spatial reasoning is quantified. Unlike purely textual tasks, spatial reasoning tasks often involve heterogeneous modalities and objectives—ranging from factual QA to embodied navigation—requiring a diverse set of evaluation protocols. Table \ref{tab:metrics} gives an overview of evaluation metrics used in spatial reasoning tasks. We group existing metrics into four thematic categories: Metrics for factual and classification task, Metrics for Language Generation, Metrics for Spatial Grounding and Geometry, Human Evaluation.

\subsubsection{Metrics for Factual and Classification Tasks}

\begin{table}[!t]
\centering
\footnotesize
\begin{tabularx}{\linewidth}{
    >{\centering\arraybackslash}X  
    >{\centering\arraybackslash}p{2.7cm}  
    >{\centering\arraybackslash}X  
    >{\centering\arraybackslash}X  
}
\toprule
\textbf{Metric} & \textbf{Formula} & \textbf{Task Type} & \textbf{Notes} \\
\midrule
Accuracy &
$\displaystyle \frac{1}{N}\sum_{i=1}^{N}\mathbb{I}(\hat{y}_i=y_i)$ &
Classification &
Measures exact correctness of discrete predictions. \\

F1 Score &
$\displaystyle F_1 = 2\cdot\frac{PR}{P+R}$ &
Classification &
Balances precision and recall. \\

BLEU\citep{papineni2002bleu} &
$\displaystyle \exp\!\left(\sum_{n} w_n \log p_n\right)$ &
Language generation &
Measures n-gram overlap with reference text. \\

ROUGE\citep{lin2004rouge} &
$\displaystyle \frac{LCS(X,Y)}{|Y|}$ &
Language generation &
Longest common subsequence ratio. \\

CIDEr\citep{vedantam2015cider} &
$\displaystyle \frac{1}{M}\sum_i \frac{g_i \cdot r_i}{\|g_i\|\|r_i\|}$ &
Language generation &
TF-IDF weighted n-gram similarity. \\

IoU &
$\displaystyle \frac{|A\cap B|}{|A\cup B|}$ &
Grounding, segmentation &
Overlap between prediction and ground truth. \\

Chamfer Dist. (CD)\citep{wu2021densityawarechamferdistancecomprehensive} &
$\displaystyle \sum_{p\in P}\min_{q\in Q}\|p-q\|^2$ &
3D generation, reconstruction &
Measures geometric closeness of point sets. \\

EMD\citep{Erickson_2021} &
$\displaystyle \min_{\phi}\sum_{p\in P}\|p-\phi(p)\|$ &
3D generation, reconstruction &
Evaluates minimal transport cost. \\

SPL\citep{yokoyama2023successweightedcompletiontime} &
$\displaystyle \frac{1}{N}\sum_i S_i \frac{L_i}{\max(P_i,L_i)}$ &
Embodied navigation &
Combines success rate and path length. \\

LLM-as-Judge\citep{li2024llms} &
N/A (LLM-based scoring model) &
Open-ended QA, captioning, reasoning &
LLMs as evaluators to assess semantic plausibility. \\

Human Evaluation &
N/A (manual scoring) &
All open-ended or generative tasks &
Human annotators assess output quality. \\
\bottomrule
\end{tabularx}
\caption{Overview of evaluation metrics used in spatial reasoning tasks.}
\label{tab:metrics}
\end{table}

For benchmarks that adopt a QA or classification format with a single ground-truth label, standard metrics such as Accuracy and F1 Score are widely used. For example, SpartQA\citep{mirzaee-etal-2021-spartqa} and STAR\citep{wu2021star_situated_reasoning} evaluate binary or multi-class answers by measuring the proportion of correct predictions. Some works also transfer other task-specific metrics into accuracy-based forms for simplicity, such as reporting results with Acc@0.5IoU in grounding tasks. While simple and interpretable, these metrics fail to capture whether the reasoning process is valid, or whether a partially correct but spatially plausible answer should be rewarded.

\subsubsection{Metrics for Language Generation}
In captioning, dialogue, or reasoning-generation settings, benchmarks often adopt BLEU\citep{papineni2002bleu}, ROUGE\citep{lin2004rouge} and CIDEr\citep{vedantam2015cider}, which quantify surface-level n-gram overlap with reference texts. Despite their popularity, these metrics exhibit poor correlation with semantic correctness, especially when evaluating fine-grained spatial relations. For example, ``The chair is left of the table and red'' has high BLEU score with the ground truth ``The chair is left of the table and green''. But they are very different in semantic meaning. 

To mitigate this limitation, recent works incorporate LLM-based judges \citep{li2024llms}, which leverage the generative and evaluative capacity of large language models to score outputs in a more semantically aligned manner. Instead of focusing solely on n-gram overlap, an LLM-based judge can assess whether a generated answer or caption is factually consistent, spatially plausible, and contextually appropriate with respect to the input scene or question. For example, in a visual question answering task, if the reference answer is “The lamp is behind the sofa” and the model predicts “The sofa is in front of the lamp”, traditional metrics like BLEU or ROUGE would penalize the surface mismatch, whereas an LLM judge can recognize that the two statements are semantically equivalent. Similarly, in captioning tasks for 3D scenes, an LLM judge can distinguish between “A red chair is to the left of the table” and “The table is on the right side of a red chair,” both of which convey the same spatial relation.  There are several recent works. GPTscore\citep{fu2023gptscoreevaluatedesire} utilizes the mergent abilities (e.g., zero-shot instruction) of LLM to score generated texts.  MAJ-EVAL\citep{chen2025multiagentasjudgealigningllmagentbasedautomated} developed a Multi-Agent-as-Judge evaluation framework that can automatically construct multiple evaluator personas with distinct backgrounds and engage in-group debates with multi-agents to generate feedbacks.

\subsubsection{Metrics for Spatial Grounding and Geometry}
Spatial reasoning uniquely requires metrics that directly assess geometric correctness. For grounding tasks, Intersection over Union(IoU) and related thresholds(Accuracy@IoU) evaluate whether predicted regions match ground-truth objects. For navigation tasks, embodied AI benchmarks adopt Success Rate (SR) and Success weighted by Path Length (SPL)\citep{yokoyama2023successweightedcompletiontime} to measure both correctness and efficiency of trajectory planning. For 3D generation and reconstruction, geometric similarity metrics such as Chamfer Distance (CD)\citep{wu2021densityawarechamferdistancecomprehensive} and Earth Mover’s Distance (EMD)\citep{Erickson_2021} are employed to capture alignment between predicted and reference shapes. These metrics are crucial as they explicitly encode spatial fidelity rather than linguistic similarity.

\subsubsection{Human Evaluation}
For the most complex and open-ended tasks, especially those involving scene-level reasoning, compositional generation, or dynamic interactions, automated metrics remain insufficient. In such cases, human-in-the-loop\citep{wu2022survey} evaluation is often the gold standard. For instance, benchmarks like Q-SpatialBench\citep{liao2024reasoningpathsreferenceobjects} and EmbSpatialBench\citep{du-etal-2024-embspatial} rely on human annotators to assess the plausibility of model outputs or the correctness of a spatial plan. These annotations are regarded the ground truth. While costly and less scalable, human evaluation provides irreplaceable insights into nuance, creativity, and commonsense plausibility—dimensions still beyond the reach of current automatic metrics.

In summary, the evaluation of spatial reasoning remains fragmented across task types: while factual metrics offer reliability, spatial reasoning demands geometry-aware measures and, in many cases, human judgment to fully capture the quality of reasoning.

\section{Methods to Improve Spatial Reasoning}
\begin{table*}[t]
\centering
\begin{adjustbox}{max width=\textwidth}

\begin{tabular}{lc|ccc|cc}
\toprule
 & \multirow{2}{*}{\textbf{\Large Methods}}
 & \multirow{2}{*}{\textbf{\Large Technique}}
 & \multirow{2}{*}{\textbf{\Large Modality}} 
 & \multirow{2}{*}{\textbf{\Large LLM Backbone}}
& \multirow{2}{*}{\textbf{\Large Tasks}}
 & \multirow{2}{*}{\textbf{\Large Cognitive Function}} \\
&  & &  &  &  &  \\
\midrule

\multirow{22}{*}{\rotatebox{90}{Training}}    
& SpatialRGPT \citep{fu2024scene} &3D encoder & Image & Qwen-VL & Spatial VQA,grounding & E-Ql-S \\

& RoboRefer \citep{zhou2025roboreferspatialreferringreasoning} &SFT, GRPO & Video, image & NVILA-2B/8B & Spatial VQA,grounding & E-Ql-S \\

& SR-3D \citep{cheng20253dawareregionprompted} &3D positional embedding & Video,image & Qwen2-7B & Spatial VQA & E-Ql-S,Qn-S \\

& SpaRE \citep{ogezi2025spareenhancingspatialreasoning} &Spatial data training & Image & Qwen2-VL & Spatial VQA & E-Ql-S, Qn-S \\

& SpaceR \citep{ogezi2025spareenhancingspatialreasoning} & RLVR, Spatial data training & Image & Qwen-2.5-VL-7B-Instruct & Spatial VQA & E-Ql-S, Qn-S \\

& MetaSpatial \citep{pan2025metaspatialreinforcing3dspatial} & RLVR & Image & Qwen2.5-VL 3B/7B & 3D generation & E-Ql-D \\

& Embodied-R \citep{zhao2025embodiedrcollaborativeframeworkactivating} &GRPO, frame selection & Video & Qwen2.5-VL, Qwen2.5 & Spatial VQA, navigation & E-Ql-S, E-Ql-D \\

& SAT \citep{ray2025satdynamicspatialaptitude} &Spatial data training & Image, video & LLaVA-1.5, LLaVA-Video & Spatial VQA,grounding & E-Ql-S,E-Ql-D \\

& ViCA2 \citep{feng2025visuospatialcognitionhierarchicalfusion} & 3D encoder & Image, video & Qwen2-7B & Spatial VQA & E-Ql-S,Qn-S \\

& Scene-LLM \citep{fu2024scene} &3D feature encoding & Image, point cloud  & LLaMA-2-7B & Spatial VQA,planning & E-Ql-S, E-Ql-D \\

& Chat-3D \citep{wang2023chat3d} &3D encoder,instruction tuning & Image, point cloud  & Vicuna-7B & Spatial VQA,caption & E-Ql-S \\

& ShapeGPT \citep{yin2025shapegpt} &3D encoder,instruction tuning & 3D mesh  & T5 & 3D generation,caption & I-Ql-S, I-Ql-D \\

& Point-LLM \citep{guo2023point} &3D encoder,PEFT & Point cloud, audio  & LLaMA & 3D generation, General QA & I-Ql-S, I-Ql-D \\

& NaviLLM \citep{zheng2024towards} &3D encoder,instruction tuning & Image  & Vicuna-7B & Spatial VQA, navigation & E-Ql-S, E-Ql-D \\

& Uni3D-LLM \citep{liu2024uni3d} & 3D encoder,PEFT  & point cloud, image  & Sphinx & Spatial VQA,3D generation & I-Ql-S, E-Ql-S \\

& ManipLLM \citep{li2024manipllm} & Task oriented tuning  & Image  & LLaMA & Object manipulation & I-Ql-D \\

& ManipLVM-R1 \citep{song2025maniplvmr1reinforcementlearningreasoning} & RLVR  & Image  & Qwen2.5-VL-3B & Object manipulation & E-Ql-D \\

& Pixel Reasoner \citep{su2025pixelreasonerincentivizingpixelspace} & RL, SFT  & Video  & Qwen2.5-VL-7B & general VQA,  & E-Ql-S, E-Ql-D \\

& 3DMIT \citep{li20243dmit} & 3D perceiver,instruction tuning  & point cloud  & Vicuna-7B, LLaVA-1.5-7B & Spatial VQA,grounding & E-Ql-S \\

& SpatialVLM \citep{Chen_2024_CVPR} & Spatial data pretraining & Image, point cloud  & PaLM-2-E & Spatial VQA, robotics & E-Ql-S, Qn-S \\

& LLaVA-3D \citep{zhu2024llava} &3D patch, 3D positional encoding & Video, 3D mesh  & LLaVA-Video-7B & general VQA ,caption & E-Ql-S, E-Ql-D \\

& ShapeLLM \citep{qi2024shapellm} &Spatial-aware encoder & Image, point cloud  & LLaMA-7B, Vicuna-13B & Spatial VQA,grounding & E-Ql-S, E-Ql-D \\

\midrule
\multirow{9}{*}{\rotatebox{90}{Inference}}  
& VoT \citep{wu2024minds} & Visualization-of-Thought & Text  & GPT-4V,Llama3 & Relational QA, navigation & E-Ql-S,E-QL-D \\

& \citep{wang2024dspybasedneuralsymbolicpipelineenhance} & DSPy & Text  & Deepseek,Llama3 & Relational QA & E-Ql-S \\

& VADAR\citep{marsili2025visualagenticaispatial} & Program synthesis & Image  & GPT-4o & Spatial QA & E-Ql-S,Qn-S \\

& $SG^2$\citep{chen2025schemaguidedscenegraphreasoningbased} & Scene graph, multi-agent & Image  & GPT-4o & Spatial QA & E-Ql-S,E-Ql-D \\

& ADAPTVIS \citep{chen2025spatialreasoninghardvlms} & Attention distribution adapt & Image & LLaVA-1.5  & Spatial VQA & E-Ql-S \\

& Agent3D-Zero \citep{zhang2024agent3dzeroagentzeroshot3d} & SoLP, view selection & Image, 3D mesh  & GPT-4V & Spatial VQA,navigation & E-Ql-S, E-Ql-D \\

& SeeGround \citep{li2025zero} & Query-aligned rendering & Image, point cloud  & Qwen2-VL,GPT-4V & 3D grounding & E-Ql-S \\

& SG-Nav \citep{yin2024sgnavonline3dscene} & Scene graph & video, image & LLaMA-7B,GPT-4 & Spatial QA, navigation & E-Ql-S, Q-Ql-D \\

& LLM-Planner \citep{song2023llm} & Few shot prompting & Image & GPT-3 & Navigation,planning & E-Ql-D \\

\midrule
\multirow{4}{*}{\rotatebox{90}{Hybrid}} 
& SpatialCoT \citep{liu2025spatialcotadvancingspatialreasoning} &3D alignment,CoT & image  & Llama3.2-Vision-11B & Navigation, manipulation & I-Ql-D, E-Ql-D \\

& Spatial-MLLM \citep{wu2025spatial} &Dual-encoder,CoT & Video  & Qwen2.5-VL-3B & Spatial VQA,grounding  & E-Ql-S, E-Ql-D \\

& MVoT \citep{li2025imaginereasoningspacemultimodal} & Visualization-of-Thought & Image, text  & Anole-7B & navigation & E-QL-D \\

& \cite{yang2023improvingvisionandlanguagereasoningspatial} & 3D encoder,spatial graph & Image  & UNIMO & general QA, caption  & E-Ql-S\\

\bottomrule
\end{tabular}
\end{adjustbox}
\caption{An overview of methods for improving spatial reasoning in Large Language Models. The methods are categorized into three main paradigms: Training-based approaches that modify model parameters, Inference-based approaches that structure the reasoning process, and Hybrid approaches that combine both. For each method, we detail its core Technique, the input Modality it operates on, its underlying LLM Backbone(for inference-based method, we show the LLM it apply to in the original paper), and the primary tasks it addresses. Crucially, we also map each method to the primary Cognitive Functions it targets, using the taxonomy introduced in Section 3. The abbreviations for the cognitive functions are as follows: E-Ql-S: Extrinsic-Qualitative-Static; I-Ql-S: Intrinsic-Qualitative-Static; Qn-S: Quantitative-Static; E-Ql-D: Extrinsic-Qualitative-Dynamic; and I-Ql-D: Intrinsic-Qualitative-Dynamic.}
\label{tab:methods}
\end{table*}

\label{methods improve}
Advancing spatial reasoning in Large Language Models (LLMs) requires methods that go beyond standard pretraining on text representation. In Table \ref{tab:methods}, we give a comprehensive overview of these methods, including their technique, primary targeted modality and tasks, LLM backbone they use and cognitive functions they improve. The central challenge arises from the text-to-world representational mismatch and the representation-level grounding problem: while the physical world is continuous and geometric, LLMs learn spatial concepts as discrete statistical patterns in language rather than as grounded physical principles. To mitigate these limitations, researchers have proposed a wide spectrum of techniques that can be broadly grouped into two paradigms.

Training-based methods embed spatial knowledge directly into model parameters by introducing architectural innovations, spatially-aware pretraining objectives, or curated data sources such as synthetic and simulated environments. In contrast, inference-based methods operate at inference time, imposing external scaffolds—such as chain-of-thought variants, scene graphs, or multimodal prompting—that guide a pre-trained model toward spatially consistent solutions without altering its parameters. Conceptually, the distinction lies in whether spatial priors are imposed internally during training or guided externally during reasoning.

This section reviews representative approaches within each paradigm, highlighting their underlying principles, design choices, and trade-offs. By examining training-based and inference-based methods, our aim is to clarify how different strategies address the linguistic–geometric gap and to identify trends that point toward more spatially capable foundation models.

\subsection{Training-based Methods}
\label{tbm}

\begin{figure}[h]   
  \centering
  \includegraphics[width=1\linewidth]{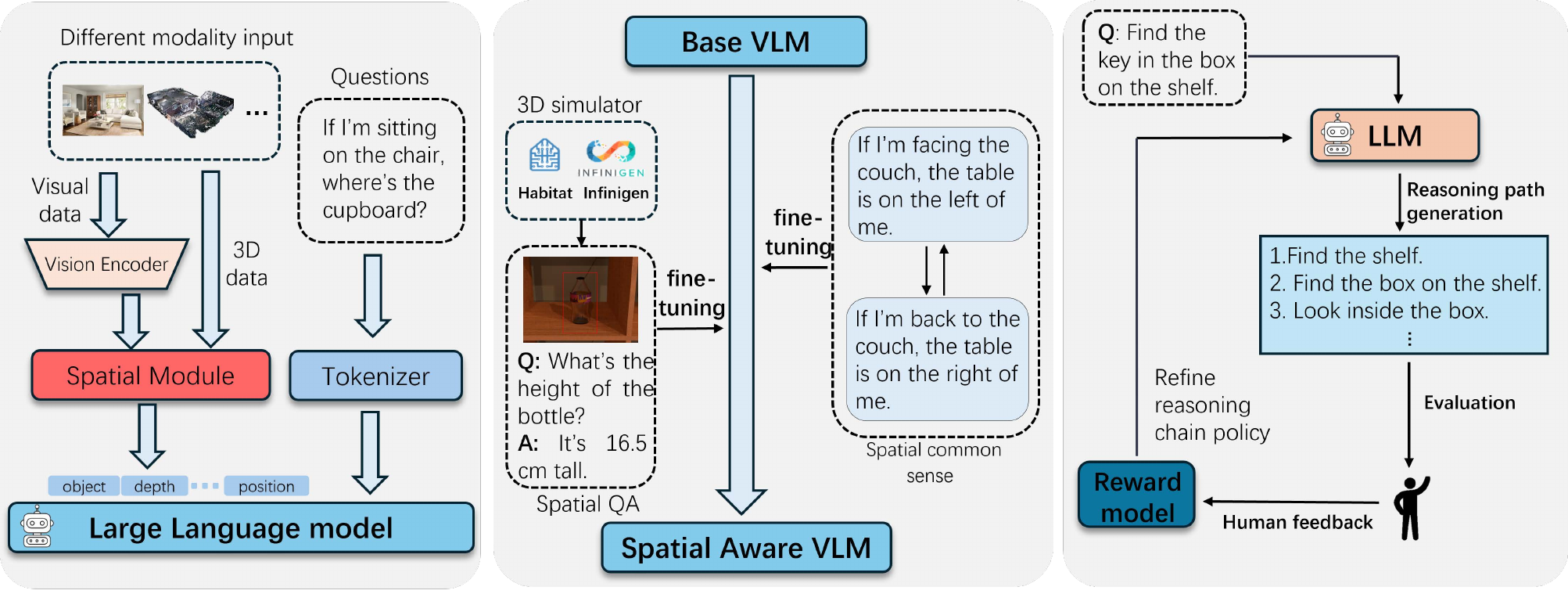}
  \caption{An overview of training-based methods. Left: Geometric priors are incorporated using spatial modules and 3D encoders to align perception with language understanding.
Middle: Synthetic environments such as Habitat or Infinigen provide controllable data for fine-tuning spatial reasoning tasks.
Right: Reinforcement learning with human feedback refines multi-step reasoning chains and improves spatial decision accuracy.}
  \label{fig:training-overview}
\end{figure}

The training-based method is the most straightforward way to endow models with spatial intelligence by directly shaping the model parameters. As shown in Figure \ref{fig:training-overview}, existing training-based methods are mainly categorized into three categories. The first category of work introduces spatial-aware modules; they train 3D encoders to encode spatial feature from images, videos, and 3D files, which explicitly capture geometric structure during representation learning. Another strategy leverages synthetic and simulation-based datasets, providing abundant, controllable examples of spatial relations that can target specific reasoning skills while mitigating annotation costs. Finally, reinforcement learning and instruction-tuning paradigms have been employed to refine reasoning paths when encountered with complex spatial reasoning questions, encouraging models to follow the explicit way of human reasoning. Together, these techniques represent the foundation of training-centric efforts to overcome the representational mismatch and build strong reasoning paradigm for large models.

\subsubsection{Spatial-Aware Module Training}
Standard Transformers are spatially naive, processing tokens as flat sequences without any intrinsic concept of 3D structure. A core approach to improving spatial reasoning is to redesign model architectures so that geometric information is explicitly embedded and aligned with the language backbone. As illustrated in Figure \ref{fig:training-overview} Left, recent works enrich visual inputs with explicit 3D priors and develop alignment layers that map these spatial features into the LLM space.

For instance, LLaVA-3D\citep{zhu2024llava} augments 2D CLIP patches with 3D positional embeddings derived from depth and camera parameters, forming 3D patches. This lets each token carry explicit spatial coordinates, enabling direct 3D reasoning without off-the-shelf 3D segmentors. Joint 2D–3D instruction tuning preserves strong 2D ability while adding robust 3D grounding and QA performance. Scene-LLM\citep{fu2024scene} constructs a hybrid point–voxel representation of egocentric and scene-level views, projecting these features into an LLM to support 3D VQA and interactive planning. PointLLM\citep{guo2023point} extends this paradigm by directly aligning colored point clouds with a pre-trained LLM through a point encoder and some projection layers. They also use Parameter-Efficient Fine-Tuning(PEFT) adapters to accelerate the alignment process. At last, SR-3D\citep{cheng20253dawareregionprompted} unifies single- and multi-view inputs by enriching 2D visual features with 3D positional embeddings derived from depth maps. It supports flexible region prompting, allowing users to annotate a region in one frame and propagate it across views for consistent spatial reasoning. This unified design leds to good performance on 3D QA and video spatial benchmarks while preserving strong 2D VLM capabilities.

The advantage of these approaches is that by injecting geometric structure directly into token representations or alignment layers, models learn spatially coherent features more efficiently, improving data efficiency and performance on tasks like 3D grounding and spatial QA. However, this specialization comes with trade-off. Additional encoders and projection layers increase architectural complexity and computational overhead, and aligning heavily to 3D priors may reduce the flexibility of LLMs as general-purpose models.

\subsubsection{Synthetic Data for Task-Specific Fine-Tuning}
A primary bottleneck in training spatially-aware models is the scarcity of large-scale, richly annotated real-world 3D data. The process of capturing, reconstructing, and manually annotating real-world scenes is expensive, time-consuming, and often results in noisy or incomplete labels. To overcome this limitation(see Figure \ref{fig:training-overview} Middle for more details), a prominent training-based strategy involves leveraging synthetic data generated from photorealistic 3D simulators and procedural generation engines like Habitat\citep{puig2023habitat3} and Infinigen\citep{infinigen2023infinite}. These environments allow for the programmatic creation of a virtually limitless number of diverse scenes, providing precise control over object placement, lighting, camera viewpoints, and physical properties.

Large quantity of data is particularly valuable for training models on specific reasoning challenges that are under-represented in real-world datasets. A model's understanding of occlusion, containment, or complex multi-object relations can be systematically improved by fine-tuning it on a synthetic dataset designed to heavily feature these scenarios. For example, SpatialVLM\citep{Chen_2024_CVPR} introduces an automatic large-scale spatial VQA data synthesis pipeline that generates 2 billion QA pairs from 10 million real-world images, infusing both qualitative and quantitative 3D relations into training. Models trained with this data gain spatial reasoning abilities including metric distance estimation and unlock new applications in chain-of-thought reasoning and robotics. SAT\citep{ray2025satdynamicspatialaptitude} generates 175K synthetic QA pairs across 22K ProcTHOR scenes. Fine-tuning LLaVA-1.5 and LLaVA-Video on SAT yields large performance gains on static benchmarks and introduces strong dynamic spatial aptitude. This demonstrates that procedurally generated synthetic data can target underrepresented reasoning skills and benefit performance in real images.

However, one common disadvantage of synthetic dataset is that they always contain some subtle patterns hidden in the synthetic process. Training on large quantity of synthetic dataset might bring about the problem of over-fitting. For example, in a generated bedroom scene, there are always 2 pillows on the 3D model of bed. That makes the LLM reply on this artificial clue to always assume there will be 2 pillows on any bed, leading to wrong results in real-world applications. As a result, researchers need to be careful when using synthetic data. Avoiding training on dataset created by similar synthetic processes may help reduce such problem.

\subsubsection{Training Reasoning Processes with Reinforcement Learning}
Beyond modifying a model’s static knowledge through architectural changes or data augmentation, a more dynamic strategy is to optimize the reasoning process itself. Complex spatial reasoning tasks often require multi-step inference that extends beyond direct perception, resembling the sequential, structured reasoning exhibited by humans. To this end, as shown in Figure \ref{fig:training-overview} Right, recent approaches increasingly frame reasoning-chain generation as a sequential decision-making problem, solvable via reinforcement learning (RL).

Inspired by Deepseek-R1\citep{deepseekai2025deepseekr1incentivizingreasoningcapability}, which introduces a verifiable, rule-guided reinforcement learning paradigm, many works adopt Group Relative Policy Optimization (GRPO) with task-specific rewards to enhance LLM for better generalization ability over pure supervised fine-tuning approach. The clipped surrogate objective of GRPO is shown below:
\begin{equation*}
J(\theta)
=
\mathbb{E}_{q,\{o_i\}}
\left[
\frac{1}{G}
\sum_{i=1}^{G}
\min\!\left(
\frac{\pi_\theta(o_i \mid q)}{\pi_{\theta_{\mathrm{old}}}(o_i \mid q)} A_i,\ 
\operatorname{clip}\!\left(
\frac{\pi_\theta(o_i \mid q)}{\pi_{\theta_{\mathrm{old}}}(o_i \mid q)},
\,1-\epsilon,\,
1+\epsilon
\right) A_i
\right)
-
\beta\, D_{\mathrm{KL}}\!\left(\pi_\theta \,\|\, \pi_{\mathrm{ref}}\right)
\right]
\end{equation*}
For each query $q$, the model samples $G$ candidate responses $\{o_i\}$ and assigns each a standardized advantage $A_i$. The objective maximizes the likelihood ratio $\frac{\pi_\theta(o_i \mid q)}{\pi_{\theta_{\mathrm{old}}}(o_i \mid q)}$ weighted by $A_i$, while applying PPO-style clipping with threshold $\epsilon$ for training stability.\citep{schulman2017proximal} A KL penalty with coefficient $\beta$ constrains the updated policy, avoiding too much shift from a reference model $\pi_{\mathrm{ref}}$. Specifically, the advantage for the $i$-th response is
\begin{equation*}
A_i = \frac{R_i - \operatorname{mean}(\{R_i\})}{\operatorname{std}(\{R_i\})},
\end{equation*}
where $R_i$ is the reward assigned to the $i$-th sample. This normalization emphasizes responses that outperform others in the same group while stabilizing optimization. The reward for each generated response is formulated as a weighted combination of format correctness, task accuracy, and length regularization, where task specific reward can be injected into the term of task accuracy:
\begin{equation*}
R = w_f R_{\mathrm{format}} + w_t R_{\mathrm{task}} + w_l R_{\mathrm{length}}.
\end{equation*}

RL based methods typically follow a two-phase paradigm. In the first phase, they adopt supervised fine-tuning(SFT) first to familiarize LLM with some predefined reasoning steps or operations. In some cases this step may be skipped because they assume the base models they use are already strong enough to conduct structured reasoning process after initial post-training. Then in the second phase, they apply reinforcement learning that improves flexibility in step selection and enhances task-specific accuracy with special designed reward functions. For example, \cite{su2025pixelreasonerincentivizingpixelspace} use instruction tuning as a warm start, exposing the LLM to expert analysis trajectories and visual cue grounding strategies. Then they use "Curiosity-Driven Reinforcement Learning" to balance exploration between pixel-space reasoning and textual reasoning. To solve spatial reasoning tasks in embodied scenario, \cite{zhao2025embodiedrcollaborativeframeworkactivating} use a specific reasoning process reward
tailored for embodied tasks as well as rule-based accuracy rewards. \cite{song2025maniplvmr1reinforcementlearningreasoning} also adopt Reinforcement Learning using Verifiable
Rewards (RLVR) by using structured, task-aligned rewards they specifically design for object manipulation tasks over two key subtasks: affordance perception and trajectory prediction.

The primary advantage of using RL is that it optimizes directly for end-task performance. It allows the model to explore reasoning paths, making it more flexible and potentially more robust than models trained with SFT alone. This is particularly powerful for complex, multi-step spatial tasks where a single correct reasoning path may not exist. However, the approach is not without significant challenges. RL training is notoriously unstable and sample inefficient, often requiring careful tuning and large amounts of interaction to converge. Moreover, reward design is a critical hurdle. as sparse or poorly specified rewards can hinder learning, and constructing dense, task-aligned reward functions for spatial reasoning is also a non-trivial problem.

\subsection{Inference-based methods}
\label{rbm}
\begin{figure}[h]   
  \centering
  \includegraphics[width=1\linewidth]{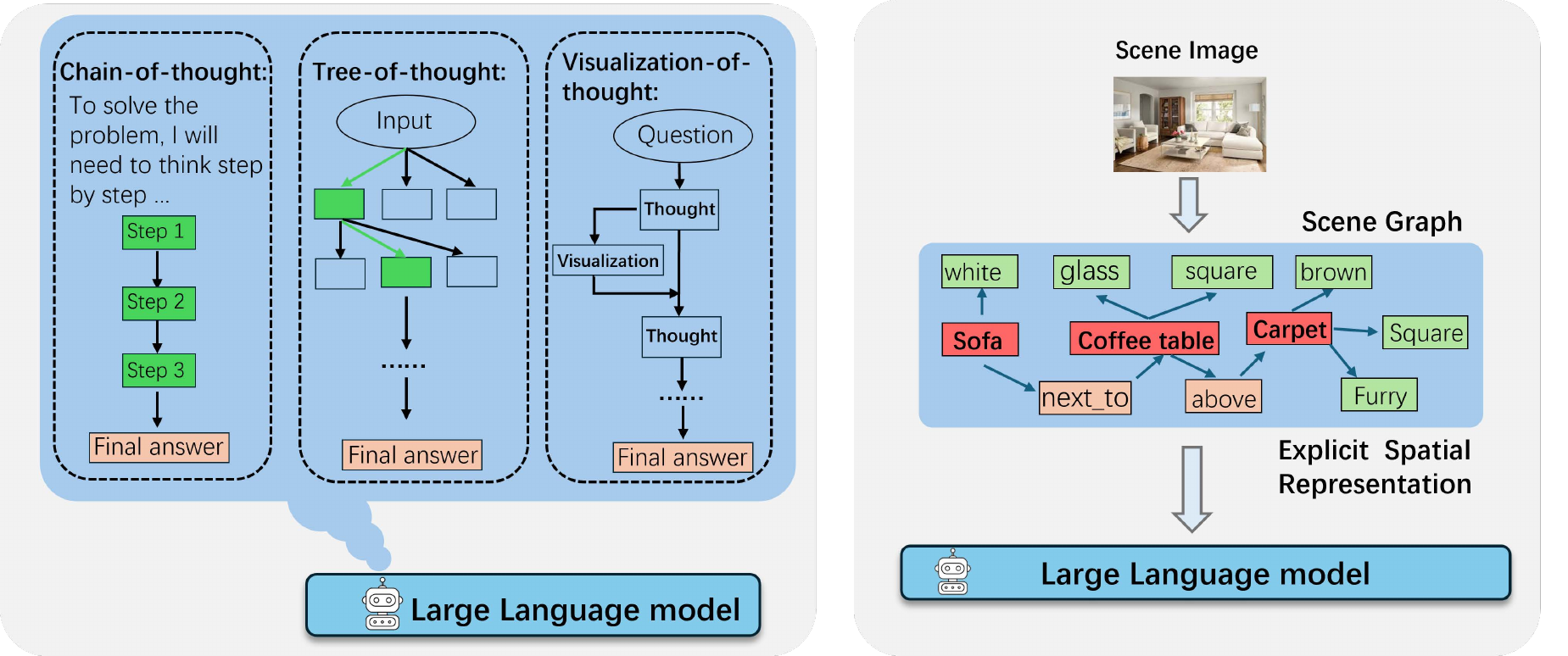}
  \caption{An overview of inference-based methods. Left: Structured prompting strategies guide models to reason step by step, branching through textual or visualized thought processes for better interpretability.
Right: Explicit spatial representations, such as scene graphs, ground linguistic reasoning in geometric relations, enabling more accurate and consistent spatial understanding.}
  \label{fig:reasoning-overview}
\end{figure}

While training-based methods aim to internalize spatial priors within model parameters, an alternative paradigm is to improve spatial reasoning at inference time through external guidance. There are mainly two directions. The first centers on chain-of-thought prompting and its variants, which structure the reasoning trajectory into interpretable steps that can better capture multi-hop relations. The second focuses on explicit spatial representations, where models construct intermediate structures such as scene graphs, cognitive maps, or grid-based layouts to ground abstract language into concrete spatial configurations. Together, these approaches highlight the potential of inference-time computing to overcome the linguistic–geometric gap without costly retraining.

\subsubsection{Chain-of-thought Prompting and Its Variants}
The most direct and widely adopted inference-based method is the use of structured prompting to guide a pre-trained model's inference process. As illustrated in Figure \ref{fig:reasoning-overview} Left, this approach is founded on the observation that while LLMs may struggle to solve complex problems in a single step, they can often succeed if prompted to break the problem down into a sequence of simpler, intermediate steps. This core technique, known as Chain-of-Thought (CoT)\citep{wei2023chainofthoughtpromptingelicitsreasoning} prompting, forces the model to externalize its reasoning process, mimicking a human-like approach to problem-solving.

The basic CoT principle has been adapted and extended in numerous ways for spatial reasoning. For instance, SpatialCoT\citep{liu2025spatialcotadvancingspatialreasoning} applies this methodology to embodied tasks like navigation and manipulation, enabling the model to perform comprehensive reasoning step by step in language space and translate this thought chain into coordinate-based actions. Similarly in spatialVLM\citep{Chen_2024_CVPR}, after pretraining the VLM on their curated dataset, they use LLM to break down complex questions into simple questions. And then they query the VLM, putting the reasoning together to derive the result.

More advanced variants move beyond simple text chains. The Visualization-of-Thought (VoT) \citep{wu2024minds} technique prompts the model to generate  ASCII-art or symbolic visualization visual intermediate steps, creating a mental sketchpad that is particularly effective for multi-step tasks like navigation and object manipulation. Multimodal Visualization-of-Thought (MVoT) \citep{li2025imaginereasoningspacemultimodal} advances this idea by enabling MLLMs to produce both verbal and dynamic visual thoughts. In this way, MVoT offers clearer interpretability and improved robustness on dynamic spatial reasoning tasks such as maze navigation and embodied planning, where traditional CoT often fails.

Despite their promise, CoT-based methods face great challenges when applied to spatial reasoning. First, their effectiveness relies on the underlying capabilities of the base model: CoT can scaffold reasoning steps, but it cannot generate fundamentally new spatial knowledge beyond what the model already learning in the training process. Second, these methods are vulnerable to error propagation, where an early hallucinated or incorrect step can cascade through subsequent reasoning, ultimately yielding a confidently incorrect final answer. Empirical evidence further underscores these challenges. For example, \cite{yang2024think} report that CoT prompting negatively impacts performance on VSI-Bench, suggesting that linear reasoning chains may even hinder models in certain spatial tasks. Consequently, the question of whether CoT truly benefits spatial reasoning remains an open and actively debated issue.

\subsubsection{Explicit Spatial Representation}
Beyond CoT-style prompting, another type of inference-based approaches guides LLMs by constructing explicit spatial representations of the scene or problem. Instead of relying solely on text tokens, these methods externalize spatial relations into structured formats such as scene graphs, cognitive maps, or logical programs, which serve as intermediate substrates for reasoning(see Figure \ref{fig:reasoning-overview} Right). This idea resonates with long-standing findings in cognitive science, where humans are believed to maintain internal cognitive map, which is a mental representations of spatial environments that support navigation and reasoning \citep{tolman1948cognitive, eichenbaum2015hippocampus, tversky1993cognitive}. The hippocampus and parietal regions are known to encode such allocentric and egocentric spatial information, forming the neural basis for spatial memory and relational reasoning. By grounding abstract queries in interpretable structures, these approaches reduce ambiguity and provide models with a clearer pathway for systematic inference.

A classical example is the use of scene graphs, which represent objects as nodes and spatial or semantic relations as edges. Scene graphs have been widely adopted in traditional CNN models for visual question answering and grounding tasks, offering a compact and relational representation of the environment\citep{Chang_2023}. And recent work\citep{yin2024sgnavonline3dscene} tries to combine it with LLMs to do zero-shot object navigation. Similarly, cognitive maps and grid-based representations have been employed, where the model maintains an internal grid map of spatial layouts to support dynamic tasks such as route planning or perspective-taking. For instance, \cite{zhang2024agent3dzeroagentzeroshot3d} introduces a novel Set-of-Line Prompting (SoLP) mechanism, overlaying Cartesian grid lines on bird’s-eye view images to provide explicit geometric references. Another line of research involves neuro-symbolic pipelines, which treat logical forms as explicit spatial representations. \cite{wang2024dspybasedneuralsymbolicpipelineenhance} introduces a DSPy-based neural-symbolic system which translates natural language descriptions into facts and rules using Answer Set Programming (ASP).

However, there are clear limitations. Constructing accurate spatial representations often requires additional modules like object detectors, parsers or symbolic solvers, introducing pipeline complexity and potential sources of error. Moreover, such graph-based methods always suffer from the same disadvantage of incompleteness as knowledge graphs. Not enough information encoded in the graph will lead to inaccurate result, while too much information brings about low efficiency in construction and searching in the graph.

\section{Open Challenges and Future Directions}
\label{open cad}
The preceding sections have illustrated that, although the integration of Multi-modal Large Language Models into visual spatial reasoning tasks has yielded notable progress\citep{li2024llavaonevisioneasyvisualtask, wang2024qwen2vlenhancingvisionlanguagemodels, chen2025expandingperformanceboundariesopensource}, the field remains in an early and uneven stage of development. Current systems demonstrate strong performance on a limited subset of challenges—particularly those involving static and qualitative scene understanding, where linguistic reasoning can be effectively anchored to perceptual inputs. However, they consistently struggle with tasks demanding metric precision, robust dynamic simulation, and compositional generalization, which reveals fundamental limitations preventing them from achieving genuine, human-like spatial intelligence. These limitations are not isolated anomalies, but systemic deficiencies spanning dataset design, training paradigms, and model architecture. Subsequently, in response to these challenges, we outline a series of promising future directions designed to guide the next phase of research toward the development of more capable and truly spatially-aware AI systems.

\subsection{Open challenges}
\label{challenges}
\noindent\textbf{Deficiencies in Datasets and Benchmarks:} A major bottleneck to progress in spatial reasoning lies in the limited availability of large-scale, high-fidelity 3D datasets, especially when compared with the web-scale corpora that power advances in language and 2D vision\citep{dai2017scannetrichlyannotated3dreconstructions}. Constructing 3D resources requires expensive capture pipelines and labor-intensive spatial annotation, leading to datasets that are typically small, domain-specific, and inconsistent in geometric detail. This scarcity constrains the diversity of spatial experiences where models can learn and improve generalization beyond synthetic settings.

Equally limiting is the imbalance in benchmark design, which has fostered a skewed impression of progress. As observed in Section 4, most existing evaluations concentrate on Extrinsic–Qualitative–Static reasoning, allowing models to achieve high accuracy through linguistic pattern matching rather than genuine geometric understanding\citep{xu2023lvlmehub}. Quantitative reasoning remains shallow, dominated by simple counting tasks rather than assessments of metric precision such as distance, angle, or scale estimation. Future benchmarks should move toward cognitively grounded evaluation, aligning task difficulty with human spatial-developmental milestones( from simple perception to mental rotation to perspective-taking) to better measure the emergence of authentic spatial competence.

\vspace{\baselineskip}

\noindent\textbf{Incomplete Spatial Understanding:} Another challenge is the incomplete and shortcut nature of spatial understanding in current models, which often demonstrates sophisticated pattern matching rather than genuine comprehension. Furthermore, this shallow understanding is also caused by a limited ability to generalize across different frames of reference. Models trained on object-centric (intrinsic) typically fail to perform environment-centered (extrinsic) transformations, indicating they have not formed a comprehensive world model. Together, these issues highlight that current systems still lack the robust, flexible spatial intelligence characteristic of human cognition.

\vspace{\baselineskip}

\noindent\textbf{Architectural and Training Paradigm Issues:}
Two deeply-rooted challenges stem from the very foundation of current models: their training paradigms and core architecture, both overwhelmingly optimized for text. The prevailing training methodology is characterized by text-dominated pre-training, where spatial grounding is a comparatively shallow fine-tuning step. This makes spatial knowledge a secondary layer on top of deeply rooted linguistic priors. Models learn to exploit statistical correlations from language rather than internalizing the underlying physical principles that govern a scene.

The aforementioned training challenges are further compounded by the Transformer's inherent design for processing one-dimensional sequences of discrete tokens, which presents a fundamental incompatibility with the continuous and volumetric structure of physical space\citep{jaegle2021perceivergeneralperceptioniterative}. This intrinsic sequence–space mismatch limits the model’s capacity to encode precise geometric relations and thereby constrains its ability to perform metric reasoning. Moreover, standard Transformer architectures are inherently stateless and lack mechanisms for maintaining persistent spatial memory. In the absence of an explicit, dynamically updatable internal model, they are unable to accumulate and integrate spatial information across time or from multiple viewpoints. Consequently, such architectural constraints severely impair performance on dynamic or embodied tasks that demand long-horizon planning, temporal consistency, or multi-step interaction\citep{ruis2020benchmarksystematicgeneralizationgrounded}. Collectively, these limitations form a self-reinforcing cycle, yielding systems that exhibit strong linguistic fluency yet remain fundamentally deficient in spatial intelligence.

\subsection{Future Directions}
\label{future directions}
\noindent\textbf{Building High Quality Datasets and Benchmarks:}
The most immediate and useful direction is the creation of richer data and more sophisticated benchmarks. The field must move beyond the current data scarcity by developing large-scale, multi-modal datasets with explicit, consistent, and rich spatial annotations. This includes not just object labels but also precise 3D coordinates, physical properties, part-whole hierarchies, and functional relationships, which are essential for deeper, physically-grounded understanding.
In parallel, we will also need to design the next generation of benchmarks that address the current evaluation gaps. This involves:
(1) A Focus on Metric Reasoning: Developing challenges that require models to estimate, compare, and reason about real-world distances, sizes, angles, and volumes, moving beyond the superficiality of object counting.
(2) Emphasizing Dynamic and Transformational Tasks: Creating more benchmarks centered on physics-based prediction, complex perspective-taking, and mental object rotation to test a model's ability to simulate and reason about a changing world.
(3) Standardizing Factor Analysis: Designing protocols that explicitly test a model's spatial reasoning ability in a quantitative way. Examine which factors(e.g. scene complexity, number of views, camera distance) possibly affect LLM's performance and try to boost in that conditions.

\vspace{\baselineskip}

\noindent\textbf{Developing Spatially-Aware Training Strategies:}
Future research should break away from the text-dominated pre-training paradigm. A promising direction lies in joint multi-modal pre-training, where language, vision, and 3D geometric representations are learned simultaneously from the ground up, rather than being layered sequentially. This approach would encourage the model to form a more unified and deeply integrated world model from the outset. Furthermore, to combat shortcut learning and improve grounding, new training objectives are needed that explicitly align representations across modalities. This includes developing novel cross-modal contrastive objectives that force the model to map textual concepts directly to their corresponding geometric configurations in a 3D space, fostering a more causal and less statistical understanding of spatial language.

\vspace{\baselineskip}

\noindent\textbf{Exploring Novel Architectures for Spatial Intelligence:}
To overcome the fundamental limitations of the standard Transformer, future work should explore architectures more naturally suited to spatial data. One promising avenue is the use of diffusion models for spatial reasoning. Their ability to operate on continuous latent spaces makes them a strong candidate for modeling distributions of plausible spatial arrangements and transformations, which is critical for generation and prediction under uncertainty.

Even more critically, the field must address the lack of persistent memory. This calls for research into architectures that incorporate an explicit and updatable mental model. This could take the form of a dynamic scene graph or a topological map that acts as a persistent memory component. Such a module would allow an embodied agent to integrate information over time and across different views, forming a coherent and enduring understanding of its environment. By offloading the burden of spatial memory to a dedicated architectural component, the LLM could then function as a high-level reasoning and planning engine that queries and interacts with this stable world model, combining the strengths of both language-based reasoning and continuous geometric representation.


\section{Conclusion}
Spatial intelligence represents one of the final frontiers in bridging language understanding with grounded physical reasoning. This survey has provided a comprehensive overview of how MLLMs approach spatial reasoning—covering cognitive foundations, task taxonomy, benchmarks, evaluation metrics and recent methodological advances. By introducing a cognitive-function-based taxonomy and a hierarchy of reasoning complexity, we have offered a principled framework that enables systematic comparison across modalities, spanning from text-only reasoning to embodied 3D interaction. This framework reveals that current progress is uneven: while models excel at static qualitative descriptions, they remain brittle in metric, dynamic, and compositional reasoning.

Through a detailed examination of datasets and benchmarks, we identified that this field is heavily shaped by synthetic and relational QA datasets, with limited coverage of quantitative or transformation-based reasoning. Our analysis of evaluation metrics further highlights a fragmentation of standards, where surface-level metrics such as accuracy or BLEU fail to evaluate geometric and cognitive understanding. Meanwhile, the review of training and reasoning-based methods illustrates two complementary paradigms: parameter-level learning to embed spatial priors, and inference-time computing guidance through structured reasoning or explicit spatial representations. Together, they demonstrate both the promise and the limits of current approaches.

Looking ahead, achieving genuine spatial intelligence will require a paradigm shift along three dimensions:
(1) Representation, developing architectures that can natively encode and manipulate 3D geometric information while retaining linguistic abstraction.
(2) Learning, integrating cross-modal and reinforcement-driven training objectives that align perception with reasoning.
(3) Evaluation, establishing cognitively grounded benchmarks that measure developmental progression—from basic perception to mental rotation and perspective-taking.

Ultimately, endowing foundation models with spatial awareness is not merely an incremental improvement but a critical step toward embodied, contextually grounded AI—systems capable of reasoning, acting, and communicating within the physical world. By synthesizing the field’s current insights and challenges, this survey aims to chart a roadmap toward that vision, encouraging the community to pursue spatial reasoning as a cornerstone of next-generation artificial intelligence.


\bibliography{main}

\bibliographystyle{tmlr}
\end{document}